\documentclass[journal, twocolumn]{IEEEtran}
\IEEEoverridecommandlockouts

\usepackage{times}    
\usepackage{amsmath}  
\usepackage{url}
\usepackage{hyperref}
\usepackage[dvips]{epsfig}
\usepackage[dvipsnames]{xcolor}
\usepackage{amsfonts}
\usepackage{amssymb,amsmath}
\usepackage{graphicx}
\usepackage{cases}
\usepackage{multirow}
\usepackage{tikz}
\usepackage{epstopdf}
\usepackage{psfrag}
\usepackage{cancel}
\usepackage{arydshln}
\usepackage{wasysym}
\usepackage{latexsym}
\usepackage{subfig}
\usepackage{lipsum}                     
\usepackage{xargs}
\usepackage{xcolor}
\hypersetup{
    colorlinks,
    linkcolor={green},
    citecolor={red},
    urlcolor={blue}
}

\usepackage{mathtools}

\usepackage[colorinlistoftodos,prependcaption,textsize=tiny]{todonotes}
\newcommandx{\unsure}[2][1=]{\todo[linecolor=red,backgroundcolor=red!25,bordercolor=red,#1]{#2}}
\newcommandx{\change}[2][1=]{\todo[linecolor=blue,backgroundcolor=blue!25,bordercolor=blue,#1]{#2}}
\newcommandx{\info}[2][1=]{\todo[linecolor=green,backgroundcolor=green!25,bordercolor=green,#1]{#2}}
\newcommandx{\improvement}[2][1=]{\todo[linecolor=Plum,backgroundcolor=Plum!25,bordercolor=Plum,#1]{#2}}
\newcommandx{\thiswillnotshow}[2][1=]{\todo[disable,#1]{#2}}

\usepackage{pdflscape}
\usepackage{afterpage}
\usepackage{soul}
\usepackage[font=scriptsize]{caption}

\title{\Large Navigating the Landscape for Real-time Localisation and Mapping for Robotics and Virtual and Augmented Reality}

\def\edinburgh{$^{\star}$}
\def\imperial{$^{\bullet}$}
\def\manchester{$^{\dagger}$}
\def\stanford{$^{\ddagger}$}
\author{
Sajad Saeedi\imperial{},
Bruno Bodin\edinburgh{},
Harry Wagstaff\edinburgh{},
Andy Nisbet\manchester{},
Luigi Nardi\stanford{},
John Mawer\manchester{},
Nicolas Melot\imperial{},\\
Oscar Palomar\manchester{},
Emanuele Vespa\imperial{},
Tom Spink\edinburgh{},  
Cosmin Gorgovan\manchester{},
Andrew Webb\manchester{},
James Clarkson\manchester{},\\
Erik Tomusk\edinburgh{},
Thomas Debrunner\imperial{},
Kuba Kaszyk \edinburgh{},
Pablo Gonzalez-de-Aledo\imperial{},
Andrey Rodchenko\manchester{},\\
Graham Riley\manchester{}, 
Christos Kotselidis\manchester{},
Bj\"{o}rn Franke\edinburgh{}, 
Michael F. P. O'Boyle\edinburgh{},  
Andrew J. Davison\imperial{},\\
Paul H. J. Kelly\imperial{}, 
Mikel Luj\'an\manchester{}, and
Steve Furber\manchester{}
   \vspace{-8 mm}
   \thanks{%
     \imperial{}~Department of Computing, Imperial College London, UK
   }
   \thanks{%
     \edinburgh{}~School of Informatics, University of Edinburgh, UK
   }
   \thanks{%
     \manchester{}~School of Computer Science, University of Manchester, UK
   }
  \thanks{%
     \stanford{}~Electrical Engineering - Computer Systems, Stanford University, USA
   }  
}

\newcommand{\tobedoneby}[1]{}
\newcommand{\completedby}[1]{}

\begin{document}
\maketitle
\begin{abstract}
Visual understanding of 3D environments in real-time, at low power, is a huge computational challenge. Often referred to as SLAM (Simultaneous Localisation and Mapping), it is central to applications spanning domestic and industrial robotics, autonomous vehicles, virtual and augmented reality.  This paper describes the results of a major research effort to assemble the algorithms, architectures, tools, and systems software needed to enable delivery of SLAM, by supporting applications specialists in selecting and configuring the appropriate algorithm and the appropriate hardware, and compilation pathway, to meet their performance, accuracy, and energy consumption goals. The major contributions we present are (1) tools and methodology for systematic quantitative evaluation of SLAM algorithms, (2) automated, machine-learning-guided exploration of the algorithmic and implementation design space with respect to multiple objectives, (3) end-to-end simulation tools to enable optimisation of heterogeneous, accelerated architectures for the specific algorithmic requirements of the various SLAM algorithmic approaches, and (4) tools for delivering, where appropriate,  accelerated, adaptive SLAM solutions in a managed, JIT-compiled, adaptive runtime context.


\end{abstract}

\section{Introduction}  


Programming increasingly heterogeneous systems for emerging application domains is an urgent challenge. One particular domain with massive potential is \emph{real-time 3D scene understanding}, poised to effect a radical transformation in the engagement between digital devices and the physical human world. In particular, visual Simultaneous Localisation and Mapping (SLAM), defined as determining the position and orientation of a moving camera in an unknown environment by processing image frames in real-time, has emerged to be an enabling technology for robotics and virtual/augmented reality applications. 

\begin{figure}[t]
\centering
\includegraphics[width = 0.9\linewidth]{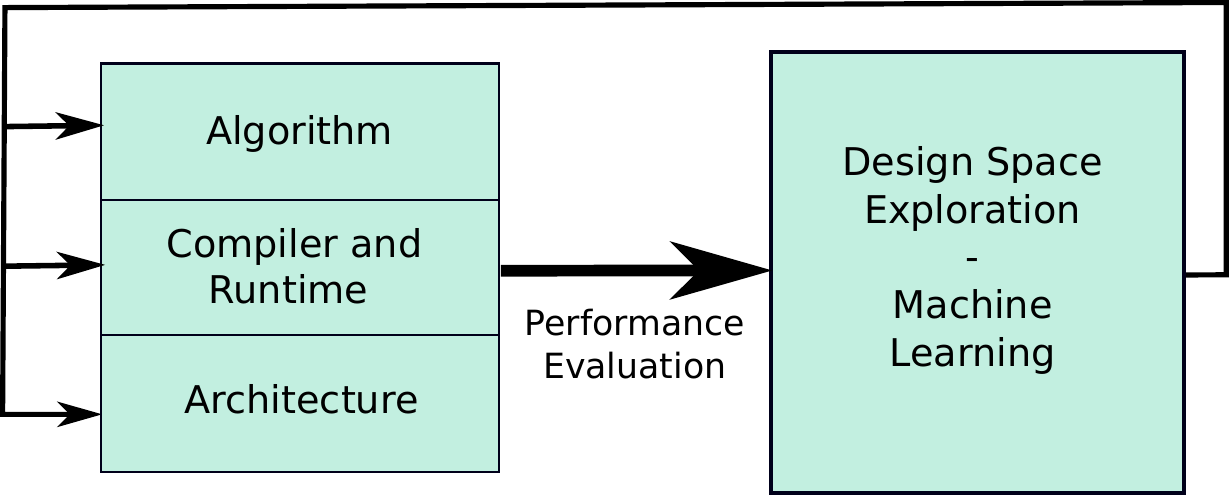}
\caption{The objective of the paper is to create a pipeline that aligns computer vision requirements with hardware capabilities. The paper's focus is on three layers: algorithms, compiler and runtime, and architecture. The goal is to develop a system that allows us to achieve power and energy efficiency, speed and runtime improvement, and accuracy/robustness at each layer and also holistically through design space exploration and machine learning techniques.}
\label{fig:pamela_overview}
\vspace{-5 mm}
\end{figure}

The objective of this work is to build the tools to enable the computer vision pipeline architecture to be designed so that SLAM requirements are aligned with hardware capability. Since SLAM is computationally very demanding, several subgoals are defined: developing systems with 1) power and energy efficiency, 2) speed and runtime improvement, and 3) improved results in terms of accuracy and robustness.
Fig.~\ref{fig:pamela_overview} presents an overview of the directions explored. At the first stage, we consider different layers of the system including architecture, compiler and runtime, and computer vision algorithms. Several distinct contributions have been presented in these three layers, explained throughout the paper. These contributions include 
novel benchmarking frameworks for SLAM algorithms, 
various scheduling techniques for software performance improvement, and 
`functional` and `detailed' hardware simulation frameworks. 
Additionally, we present holistic optimisation techniques, such as Design Space Exploration (DSE), that allows us to take into account all these layers together and optimise the system holistically to achieve the desired performance metrics.



The major contributions we present are:
\begin{itemize}
\item tools and methodology for systematic quantitative evaluation of SLAM algorithms,
\item automated, machine-learning-guided exploration of the algorithmic and implementation design space with respect to multiple objectives,
\item end-to-end simulation tools to enable optimisation of heterogeneous, accelerated architectures for the specific algorithmic requirements of the various SLAM algorithmic approaches, and 
\item tools for delivering, where appropriate,  accelerated, adaptive SLAM solutions in a managed, JIT-compiled, adaptive runtime context.
\end{itemize}
This article is an overview of a large body of work unified by these common objectives --- to apply software synthesis, and automatic performance tuning in the context of compilers and library generators, performance engineering, program generation, languages, and hardware synthesis. We specifically target mobile, embedded, and wearable contexts, where trading off quality-of-result against energy consumption is of critical importance.  The key significance of the work lies, we believe, in showing the importance and the feasibility of extending these ideas across the full stack, incorporating algorithm selection and configuration into the design space along with code generation and hardware levels of the system.”

\subsection{Background}  
Based on the structure shown in Fig.~\ref{fig:pamela_overview}, in this section, background material for the following topics is presented very briefly:
\begin{itemize}
\item computer vision, 
\item system software, 
\item computer architecture, and 
\item model-based design space exploration. 
\end{itemize}

\subsubsection{Computer Vision}
In computer vision and robotics community, SLAM is a well-known problem. Using SLAM, a sensor, such as a camera, is able to localise itself in an unknown environment by incrementally building a map and at the same time localising itself within the map. Various methods have been proposed to solve the SLAM problem, but robustness and real-time performance is still challenging~\cite{Cadena:TRO:2017}. From the mid 1990s onwards, a strong return has been made to a model-based paradigm enabled primarily by the adoption of probabilistic algorithms~\cite{Thrun:2005:PR} which are able to cope with the uncertainty in all real sensor measurements~\cite{Durrant2006:RAM}. A breakthrough was when it was shown to be feasible using computer vision applied to commodity camera hardware. The MonoSLAM system  offered real-time 3D tracking of the position of a hand-held or robot-mounted camera while reconstructing a sparse point cloud of scene landmarks~\cite{Davison2007:PAMI}. Increasing computer power has since enabled previously ``off-line'' vision techniques to be brought into the real-time domain; Parallel Tracking and Mapping (PTAM) made use of classical bundle adjustment within a real-time loop~\cite{klein09ISMAR}. 
Then live dense reconstruction methods, Dense Tracking and Mapping (DTAM) using a standard single camera~\cite{Newcombe:2011:ICCV} and KinectFusion using a Microsoft Kinect depth camera~\cite{2011Newcombe}, showed that surface reconstruction can be a standard part of a live SLAM pipeline, making use of GPU-accelerated techniques for rapid dense reconstruction and tracking. 

KinectFusion is an important research contribution and has been used throughout this paper in several sections, including in SLAMBench benchmarking (Section~\ref{sec:bench}), in improved mapping and path planning in robotic applications (Section~\ref{sec:ofusion}), in Diplomat static scheduling (Section~\ref{sssec:diplomat}), in Tornado and MaxineVM dynamic scheduling (Sections~\ref{sec:tornado} and~\ref{sec:maxinevm}), in MaxSim hardware profiling (Section~\ref{sec:maxsim}), and various design space exploration and crowdsourcing methods (Section~\ref{sec:holistic}). 

KinectFusion models the occupied space only and tells nothing about the free space which is vital for robot navigation. In this paper, we present a method to extend KinectFusion to model free space as well (Section~\ref{sec:ofusion}). Additionally, we introduce two benchmarking frameworks, SLAMBench and SLAMBench2 (Section~\ref{sec:bench}). These frameworks allow us to study various SLAM algorithms, including KinectFusion, under different hardware and software configurations. Moreover, a new sensor technology, focal-plane sensor-processor arrays, is used to develop scene understanding algorithms, operating at very high frame rates with very low power consumption (Section~\ref{sec:cpa}).


\subsubsection{System Software}

Smart scheduling strategies can bring significant performance improvement regarding  execution time~\cite{fan12} or energy consumption~\cite{melot15,melot15-parco,xu12} by breaking an algorithm into smaller units, distributing the units between cores or  Intellectual Properties (IP)s available, and adjusting the voltage and frequency of the cores. Scheduling can be done either statically or dynamically. Static scheduling requires extended knowledge about the application, i.e., how an algorithm can be broken into units, and how these units behave in different settings. 
Decomposing an algorithm this way impacts a static scheduler's choice in allocating and mapping resources to computation units, and therefore it needs to be optimised. In this paper, two static scheduling techniques are introduced (Section~\ref{sec:staticscheduling}) including idiom-based compilation and Diplomat, a task-graph framework that exploits static dataflow analysis to perform CPU/GPU mapping. 

Since static schedulers do not operate online, optimisation time is not a primary concern. However, important optimisation opportunities may depend on the \emph{data} being processed; therefore, dynamic schedulers have more chances in obtaining the best performance. In this paper, two novel dynamic scheduling techniques are introduced including 
MaxineVM, a research platform for managed runtime languages executing on ARMv7, 
and 
Tornado, a heterogeneous task-parallel programming framework designed for heterogeneous systems where the specific configurations of CPUs, GPGPUs, FPGAs, DSPs, etc. in a system are not known till runtime  (Section~\ref{sec:dynamicscheduling}).

In contrast, dynamic schedulers cannot spend too much processing power to find good solutions, as the performance penalty may outweight the benefits they bring. Quasi-static scheduling is a compromising approach that statically computes a good schedule and further improves it online depending on runtime conditions~\cite{schwarzer15}. A hybrid scheduling technique is introduced called power-aware code generation,  which is a compiler-based approach to runtime power management for heterogeneous cores (Section~\ref{sec:hybridscheduling}).


\subsubsection{Computer Architecture}

It has been shown that moving to a dynamic heterogeneous model,
where the use of hardware resources and the capabilities of those resources are adjusted at run-time, allows far more flexible optimisation of system performance and efficiency~\cite{dastgeeretal15,dastgeeretal16}. 
Simulation methods, such as memory and instruction set simulation, are powerful tools to design and evaluate such systems. A large number of simulation tools are available~\cite{Bohm:2011:GJT}; in this paper we further improve upon current tools by introducing novel `functional' and `detailed' hardware simulation packages, that can simulate individual cores and also complete CPU/GPU systems (Section~\ref{sec:simulation}). Also novel profiling (Section~\ref{sec:profiling}) and specialisation (Section~\ref{sec:specialisation}) techniques are introduced which allow us to custom-design chips for SLAM and computer vision applications.


\subsubsection{Model-based Design Space Exploration}
Machine learning has rapidly emerged as a viable means 
to automate sequential optimising compiler construction. Rather than hand-craft a set of optimisation heuristics based on compiler expert insight, learning techniques automatically determine how to apply optimisations based on statistical modelling and learning. Its great advantage is that it can adapt to changing platforms as it has no \emph{a priori} assumptions about their behaviour. There are many studies showing it outperforms human-based approaches~\cite{Cooper:2005:ACM},~\cite{Fursin:IJPP:2011},~\cite{wang:ACM:2012}, and~\cite{kulkarni:ACM:2012}.

Recent work shows that machine learning can automatically port across architecture spaces with no additional learning time, 
and can find different, appropriate, ways of mapping program parallelism for different parallel platforms~\cite{Leather:2009:IEEECom},~\cite{Tournavitis:2009:ACM}. There is now ample evidence from previous research, that design space exploration based on machine learning provides a powerful tool for optimising the configuration of complex systems both statically and dynamically. It has been used from the perspective of single-core processor design~\cite{Zuluaga"2012:DATE}, the modelling and prediction of processor performance~\cite{Bohm_2010}, the dynamic reconfiguration of memory systems for energy efficiency~\cite{Sundararajan:IEEE}, the design of SoC interconnect architectures~\cite{Almer:2011:LAA}, and power management~\cite{Sundararajan:IEEE}. 
The DSE methodology will address this paper's goals from the perspective of future many-core systems, extending beyond compilers and architecture to elements of the system stack including application choices and run-time policies. In this paper, several DSE related works are introduced. Multi-domain DSE performs exploration on hardware, software, and algorithmic choices (Section~\ref{sec:multidse}). With multi-domain DSE, it is possible to compromise between metrics such as runtime speed, power consumption, and SLAM accuracy. In Motion-aware DSE (Section~\ref{sec:msdse}), we develop a comprehensive DSE that also takes into account the complexity of the environment being modelled, including the photometric texture of the environment, the geometric shape of the environment, and the speed of the camera in the environment. DSE works allow us to design applications that can optimally choose a set of hardware, software, and algorithmic parameters meeting certain desired performance metrics. One example application is active SLAM (Section~\ref{sec:activeslam}).

\subsection{Outline}
\begin{figure}[t]
\centering
\includegraphics[width = 0.9\linewidth]{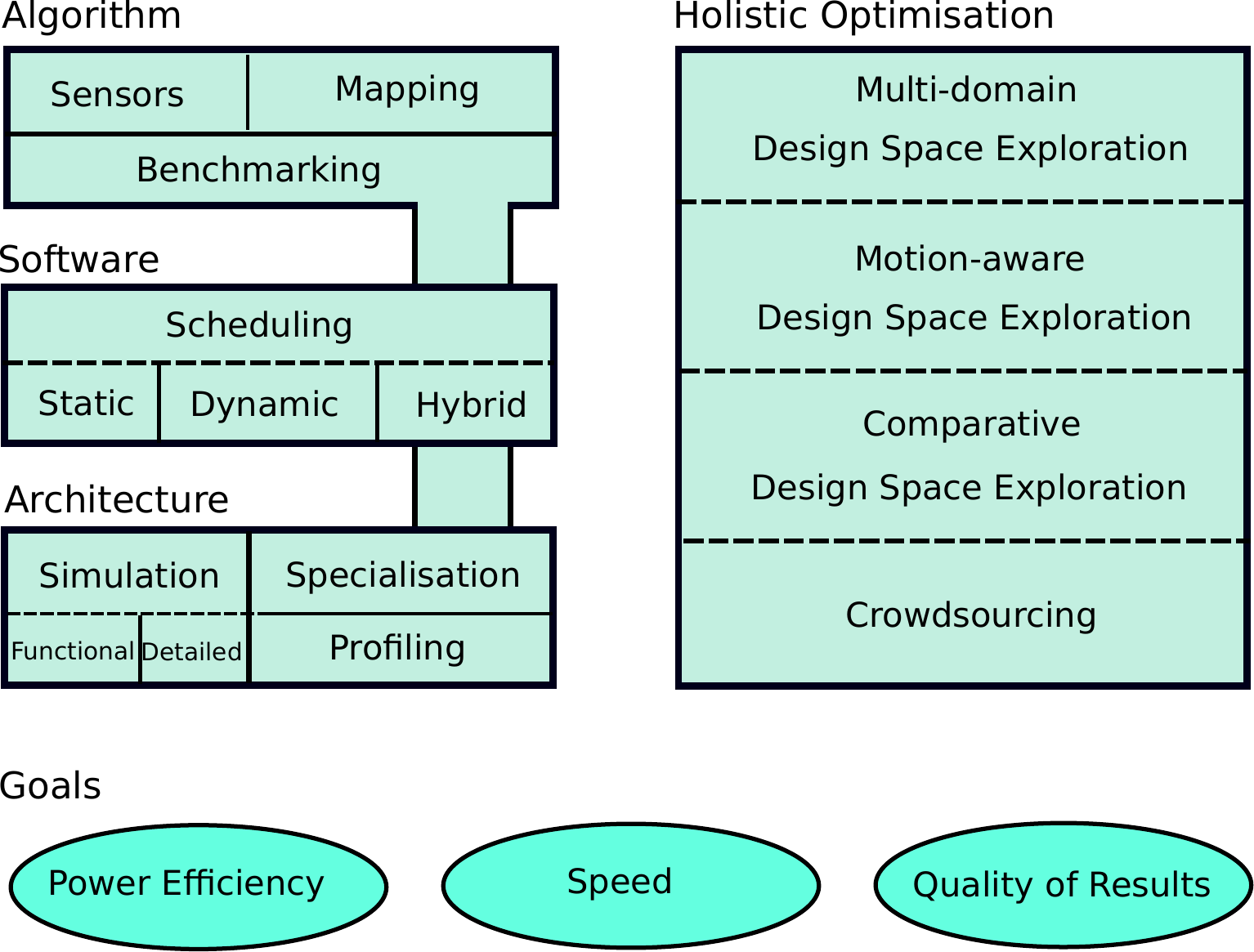}
\caption{
Outline of the paper. The contributions of the paper have been organised under four sections, shown with solid blocks. These blocks cover algorithmic, software, architecture, and holistic optimisation domains. Power efficiency, runtime speed, and quality of results are the subgoals of the project. The latter includes metrics such as accuracy of model reconstruction, accuracy of trajectory, and robustness.}
\label{fig:outline}
\end{figure}

Real-time 3D scene understanding is the main driving force behind this work. 3D scene understanding has various applications in wearable devices, mobile systems, personal assistant devices, Internet of Things, and many more. 
Throughout this paper, we aim to answer the following questions: 
1) How can we improve 3D scene understanding (specially SLAM) algorithms? 
2) How can we improve power performance for heterogeneous systems? 
3) How can we reduce the development complexity of hardware and software? 
As shown in Fig.~\ref{fig:outline}, we focus on four design domains: computer vision algorithms, software, hardware, and holistic optimisation methods. Several novel improvements have been introduced, organised as shown in Fig.~\ref{fig:outline}.
\begin{itemize}
\item Section~\ref{sec:application_algorithm} (Algorithm) explains the algorithmic contributions such as using novel sensors, improving dense mapping, and novel benchmarking methods.
\item Section~\ref{sec:software} (Software) introduces software techniques for improving system performance, including various types of scheduling.
\item Section~\ref{sec:hardware} (Architecture) presents hardware developments, including simulation, specialisation, and profiling techniques.
\item Section~\ref{sec:holistic} (Holistic Optimisation) introduces holistic optimisation approaches, such as design space exploration and crowdsourcing.
\item Section~\ref{sec:conclusions} summarises the work. 
\end{itemize}



\section{Computer Vision Algorithms and Applications} \label{sec:application_algorithm}
Computer vision algorithms are the main motivation of the paper. We focus mainly on SLAM. Within the past few decades, researchers have developed various SLAM algorithms, but few tools are available to compare and 
benchmark these algorithms and evaluate their performance on the available diverse hardware platforms. 
Moreover, the general research direction is also moving towards making the current algorithms more 
robust to eventually make them available in industries and our everyday life. Additionally, as the 
sensing technologies progress, the pool of SLAM algorithms become more diverse and fundamentally new approaches need to be invented. 

\begin{figure}[t!]
\centering
\includegraphics[width = .64\linewidth]{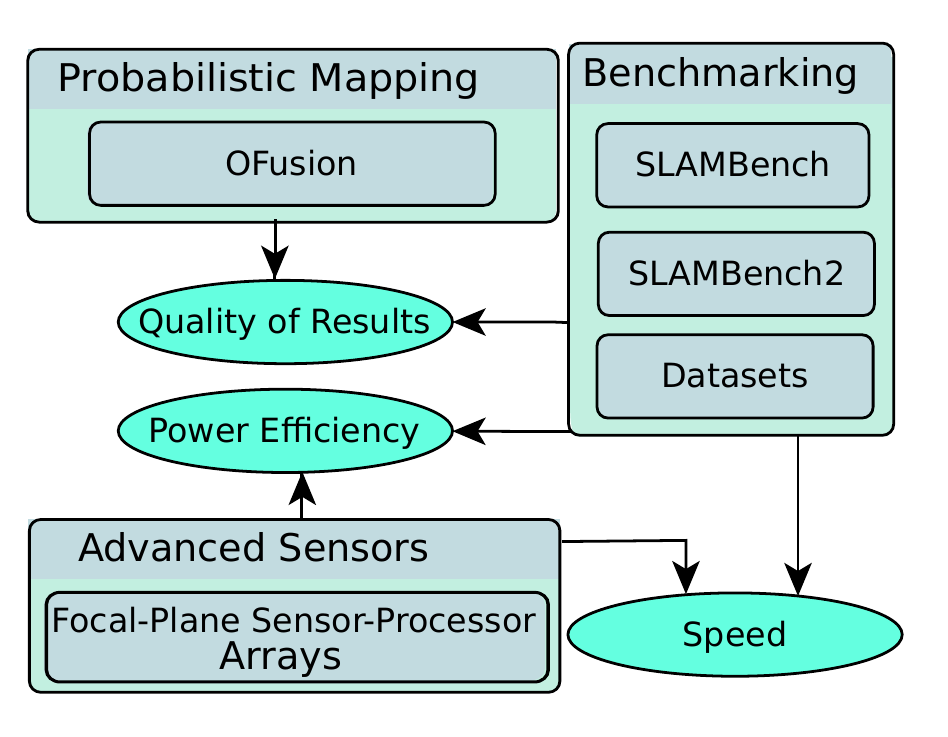}
\caption{Algorithmic contributions include benchmarking tools, advanced sensors, and improved probabilistic mapping.}
\label{fig:cv_alg}
\end{figure}

This section presents algorithmic contributions from three different aspects. As shown in Fig.~\ref{fig:cv_alg}, three main topics are covered: 1) benchmarking tools to compare the performance of the SLAM algorithms, 2) improved probabilistic mapping, and 3) new sensor technologies for scene understanding. 

\subsection{Benchmarking: Evaluation of SLAM Algorithms}\label{sec:bench}
Real-time computer vision and SLAM offer great potential for a new level of scene modelling, tracking, and real environmental interaction for many types of robots, but their high computational requirements mean that implementation on mass market embedded platforms is challenging. Meanwhile, trends in low-cost, low-power processing are towards massive parallelism and heterogeneity, making it difficult for robotics and vision researchers to implement their algorithms in a performance-portable way. 

To tackle the aforementioned challenges, in this section, two computer vision benchmarking frameworks are introduced: SLAMBench and SLAMBench2. 
Benchmarking is a scientific method to compare the performance of different hardware and software systems. 
Both benchmarking frameworks share common functionalities, but their objectives are different. While SLAMBench provides a framework that is able to benchmark various implementations of KinectFusion, 
SLAMBench2 provides a framework that is able to benchmark various different SLAM algorithms in their original implementations.

Additionally, to systemically choose the proper datasets to evaluate the SLAM algorithms, we introduce a dataset complexity scoring method. All these projects allow us to optimise power, speed, and accuracy.

\subsubsection{SLAMBench}  

\begin{figure}[t!]
\centering
\includegraphics[width =.75 \linewidth]{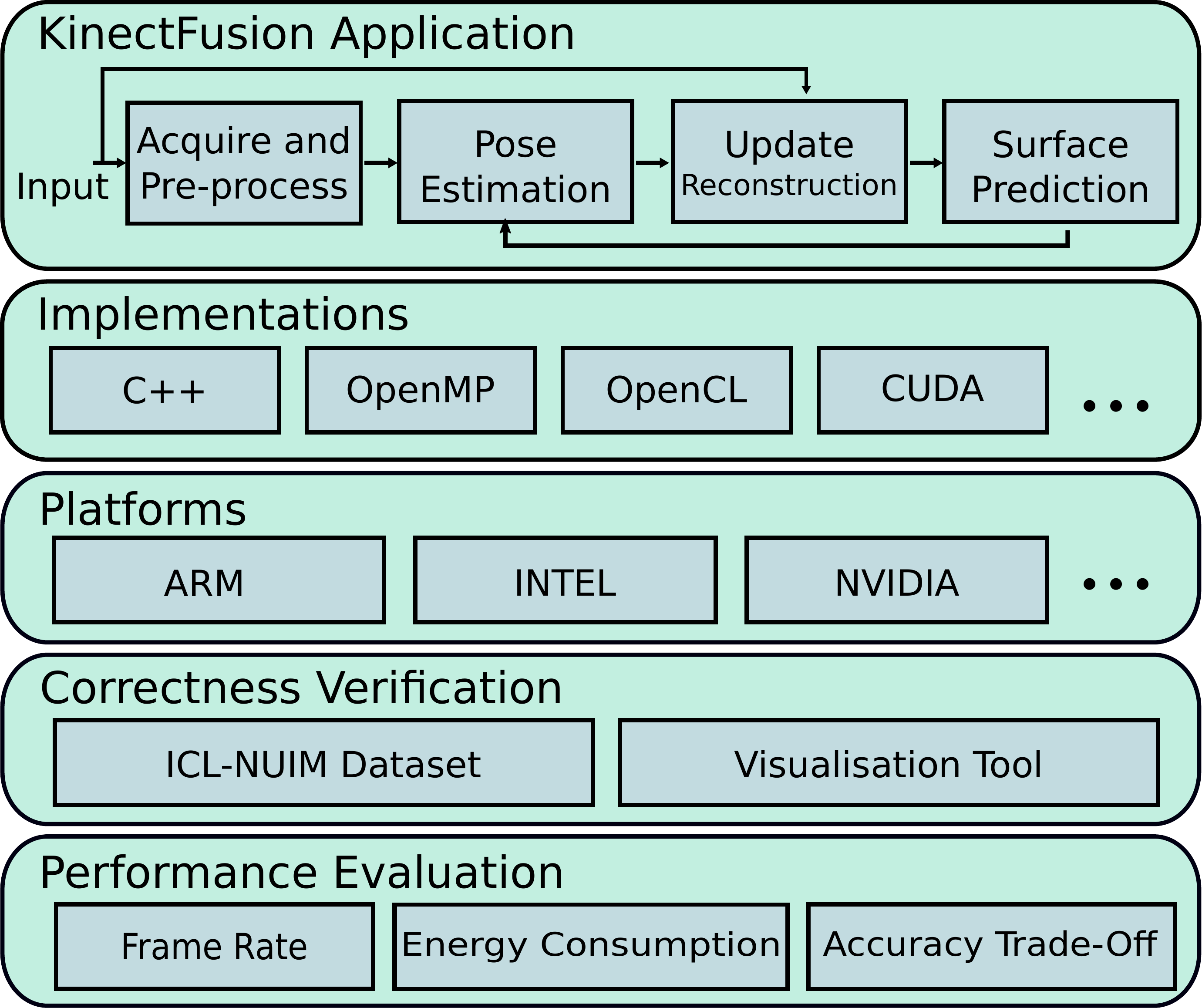}
\caption{SLAMBench enables benchmarking of the KinectFusion algorithm on various types of platforms by providing different implementations such as C++, OpenMP, CUDA, and OpenCL.
}
\label{fig:slambench_framework}
\end{figure}

As a first approach to investigate SLAM algorithms, we introduced SLAMBench~\cite{nardi2015introducing}, a publicly available software framework which represents a starting point for quantitative, comparable, and validatable experimental research to investigate trade-offs in performance, accuracy, and energy consumption of a dense RGB-D SLAM system.  
SLAMBench provides a KinectFusion~\cite{2011Newcombe} implementation, inspired by the open-source KFusion implementation~\cite{KFusion}. SLAMBench provides the same KinectFusion in the C++, OpenMP, CUDA, and OpenCL variants, and harnesses the ICL-NUIM synthetic RGB-D dataset~\cite{2014Handa} with trajectory and scene ground truth for reliable accuracy comparison of different implementation and algorithms. 
The overall vision of the SLAMBench framework is shown in Fig.~\ref{fig:slambench_framework}, refer to~\cite{nardi2015introducing} for more information.

Third parties have provided implementations of SLAMBench in additional emerging languages: 
\begin{itemize}
\item the C++ SyCL for OpenCL Khronos Group standard~\cite{keir2016dagr},
\item the platform-neutral compute intermediate language for accelerator programming PENCIL~\cite{baghdadi2015pencil}, the PENCIL SLAMBench implementation can be found in~\cite{pencil-SLAMBench}.
\end{itemize}  

As demonstrated in Fig.~\ref{fig:outline}, SLAMBench has enabled us to do more research in algorithmic, software, and architecture domains, explained throughout the paper. Examples include 
Diplomat static scheduling (Section~\ref{sssec:diplomat}), 
Tornado dynamic scheduling (Sections~\ref{sec:tornado}), 
MaxSim hardware profiling~(Section~\ref{sec:maxsim}),  
multi-domain design space exploration (Section~\ref{sec:multidse}),
comparative design space exploration (Section~\ref{sec:comparativedse}), 
and crowdsourcing (Section~\ref{sec:crowdsourcing}).

\subsubsection{SLAMBench2}   \completedby{Edinburgh - Bruno}

\begin{figure}[t!]
	\centering
	\includegraphics[width =.5 \linewidth]{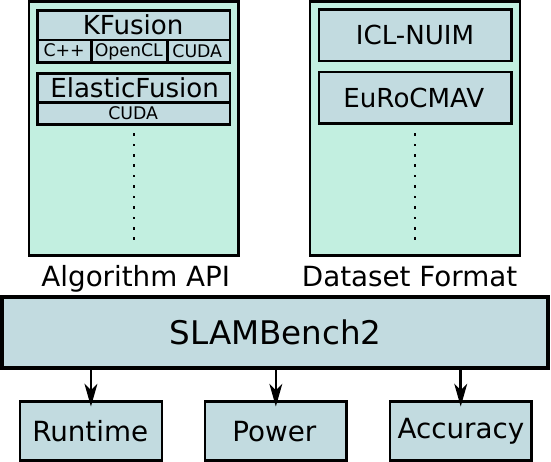}
	\caption{SLAMBench2 allows multiple algorithms (and implementations) to be combined with a wide array of datasets. A simple API and dataset make it easy to interface with new algorithms.}
    \label{fig:sb2}
\end{figure}

SLAMBench has had substantial success within both the compiler and architecture realms of academia and industry.
The SLAMBench performance evaluation framework is tailored for the KinectFusion algorithm and the ICL-NUIM input dataset. 
However, in SLAMBench 2.0, we re-engineered SLAMBench to have more modularity by integrating two major features~\cite{Bodin2018ICRA}. Firstly, a SLAM API has been defined, which provides an easy interface to integrate any new SLAM algorithms into the framework. Secondly, there is now an I/O system in SLAMBench2 which enables the easy integration of new datasets and new sensors (see Fig.~\ref{fig:sb2}). Additionally, SLAMBench2 features a new set of algorithms and datasets from among the most popular in the computer vision community, Table~\ref{tb:sb2} summarises these algorithms.


\begin{table}
	\centering
	\begin{tabular}{l l l}
		\hline
        \textbf{Algorithm} & \textbf{Type} & \textbf{Implementations} \\ 
		\hline
		ElasticFusion~\cite{Whelan_RSS_2015} & Dense & CUDA \\
		InfiniTAM~\cite{InfiniTAM_ISMAR_2015} & Dense & C++, OpenMP, CUDA \\
		KinectFusion~\cite{2011Newcombe} & Dense & C++, OpenMP, OpenCL, CUDA\\
		\hline
		LSD-SLAM~\cite{engel2014lsd} & Semi-Dense  & C++, PThread \\
		\hline
		ORB-SLAM2~\cite{murartal2016} & Sparse &   C++  \\        
		MonoSLAM~\cite{davison2007monoslam} & Sparse & C++, OpenCL \\
		OKVIS~\cite{leutenegger2015keyframe} & Sparse  & C++ \\
		PTAM~\cite{klein09ISMAR} & Sparse  & C++ \\
                SVO~\cite{Forster2014ICRA} & Sparse  & C++ \\
        \hline
	\end{tabular}
	\caption{List of SLAM algorithms currently integrated in SLAMBench2. These algorithms provide either dense, semi-dense, or sparse reconstructions~\cite{Bodin2018ICRA}.}
        \label{tb:sb2}
\end{table}

The works in~\cite{Abouzahir_RAS_2014} and~\cite{Delmerico_ICRA_2018} present benchmarking results, comparing several SLAM algorithms on various hardware platforms; however, SLAMBench2 provides a framework that researchers can easily integrate and use to explore various SLAM algorithms.

\subsubsection{Datasets} 

\completedby{Imperial - Sajad}

Research papers on SLAM often report performance metrics such as pose estimation accuracy, scene reconstruction error, or energy consumption. The reported performance metrics, may not be representative of how well an algorithm will work in real-world applications. Additionally, as the diversity of the datasets is growing, it becomes a challenging issue to decide which and how many datasets should be used to compare the results. To address this concern, not only we categorised datasets  according to their complexity in terms of trajectory and environment, but also we have proposed new synthetic datasets with highly detailed scene and realistic trajectories~\cite{Li_BMVC_2018, ICLdataset}.

In general, datasets do not come with a measure of complexity level, and thus the comparisons may not reveal all strengths or weaknesses of a new SLAM algorithm. In~\cite{Saeedi_ICRA_2017}, we proposed to use frame-by-frame Kullback-Leibler divergence as a simple and fast metric to measure the complexity of a dataset. Across all frames in a dataset, mean divergence and the variance of divergence were used to assess the complexity. Table~\ref{tb:difficulty} shows some of these statistics for ICL-NUIM sequences for intensity divergence. Based on the reported trajectory error metrics of the ElasticFusion algorithm~\cite{Whelan_RSS_2015}, datasets lr\_kt2 and lr\_kt3 are more difficult than lr\_kt0 and lr\_kt1. Using the proposed statistical divergence, these difficult trajectories have a higher complexity score as well.

\subsection{OFusion: Probabilistic Mapping}\label{sec:ofusion}

\begin{table}
\begin{center}
\begin{tabular}{ c c c c c } 
\hline
\textbf{Dataset} & \textbf{Trajectory} & \textbf{Max}  & \textbf{Mean} & \textbf{Variance}  \\
\hline
\multirow{3}{2em}{ICL-NUIM} 
& lr\_kt0 & 0.0250 & 0.0026 & 0.0014  \\ 
& lr\_kt1 & 0.0183 & 0.0026 & 0.0012 \\ 
& lr\_kt2 & 0.0427 & 0.0032 & 0.0023\\ 
& lr\_kt3 & 0.0352 & 0.0032 & 0.0023\\ 
\hline
\end{tabular}
\caption{Complexity level metrics using information divergence~\cite{Saeedi_ICRA_2017}. 
}\label{tb:difficulty}
\end{center}
\end{table}

Modern dense volumetric methods based on signed distance functions such as DTAM~\cite{Newcombe:2011:ICCV} or explicit point clouds, such as ElasticFusion~\cite{Whelan_RSS_2015}, are
able to recover high quality geometric information in real-time. However, they
do not explicitly encode information about empty space which essentially
becomes equivalent to unmapped space. In various robotic applications this
could be a problem as many navigation algorithms require explicit and
persistent knowledge about the mapped empty space. Such information is instead
well encoded in classic occupancy grids, which, on the other hand, lack the
ability to faithfully represent the surface boundaries. Loop et al.~\cite{Loop_3DV_2016} proposed a novel probabilistic fusion framework aiming at closing such information gap by employing a continuous occupancy map representation in which the surface boundaries are well-defined. Targeting real-time robotics applications, we have extended such framework to make it suitable for the incremental tracking and mapping typical of an exploratory SLAM system. The new formulation, denoted as OFusion~\cite{Vespa:ICRA:2018}, allows robots to seamlessly perform camera tracking, occupancy grid mapping and surface reconstruction at the same time. As shown in Table~\ref{tab:ate-comp}, OFusion not only encodes the free space, but also performs at the same level or better than state-of-the-art volumetric SLAM pipelines such as KinectFusion~\cite{2011Newcombe} and InfiniTAM~\cite{InfiniTAM_ISMAR_2015} in terms of mean Absolute Trajectory Error (ATE). To demonstrate the effectiveness of our approach we implemented
a simple path planning application on top of our mapping pipeline. We used
Informed RTT*~\cite{Gammell:IROS:2014} to generate a collision-free 3-meter long trajectory between two obstructed start-goal endpoints, showing the feasibility to achieve tracking,
mapping and planning in a single integrated control loop in real-time.

\begin{table}
  \center
\begin{tabular}{l c c c }
  \hline
  \textbf{Trajectory}          & \textbf{TSDF}     & \textbf{OFusion}    & \textbf{InfiniTAM} \\
  \hline
  ICL-NUIM lr\_kt0       & 0.0113   & 0.2289   & 0.3052 \\
  ICL-NUIM lr\_kt1       & 0.0117   & 0.0170   & 0.0214 \\
  ICL-NUIM lr\_kt2       & 0.0040   & 0.0055   & 0.1725 \\
  ICL-NUIM lr\_kt3       & 0.7582   & 0.0904   & 0.4858 \\
  TUM fr1\_xyz    & 0.0295   & 0.0322   & 0.0273  \\
  TUM fr1\_floor  & $\times$ & $\times$ & $\times$             \\
  TUM fr1\_plant  & $\times$ & $\times$ & $\times$             \\
  TUM fr1\_desk   & 0.1030   & 0.0918 & 0.0647        \\
  TUM fr2\_desk   & 0.0641   & 0.0724 & 0.0598        \\
  TUM fr3\_office   & 0.0686 & 0.0531   & 0.0996      \\
  \hline
\end{tabular}
  \caption{Absolute Trajectory Error (ATE), in metres, comparison between KinectFusion (TSDF), 
  occupancy mapping (OFusion), and InfiniTAM across sequences from the ICL-NUIM and 
  TUM RGB-D detasets. Cross signs indicate tracking failure. }
  \label{tab:ate-comp}
\end{table}


\subsection{Advanced Sensors}\label{sec:cpa}

\completedby{Imperial - Thomas, Sajad}



Mobile robotics and various applications of SLAM, Convolutional Neural Networks (CNN), and VR/AR are constrained by power resources and low frame rates. These applications can not only benefit from high frame rate, but also could save resources if they consumed less energy. 

Monocular cameras have been used in many scene understanding and SLAM algorithms~\cite{davison2007monoslam}. Passive stereo cameras (e.g. Bumblebee2, 48~FPS @ 2.5~W~\cite{Bumblebee2}),  structured light cameras (e,g, Kinect, 30~FPS @ 2.25~W~\cite{Fankhauser_ICRA_2015}) and Time-of-flight cameras (e.g. Kinect One, 30~FPS @ 15~W~\cite{Fankhauser_ICRA_2015}) additionally provide metric depth measurements; however, these cameras are limited by low frame rate and have relatively demanding power budget for mobile devices; problems that modern bio-inspired and analogue methods are trying to address. 

Dynamic Vision Sensor (DVS), also known as the event camera, is a novel bio-inspired imaging technology, which has the potential to address some of the key limitations of conventional imaging systems. Instead of capturing and sending a full frame, an event camera captures and sends a set of sparse \emph{events}, generated by the change in the intensity. They are low-power and are able to detect changes very quickly. Event cameras have been used in camera tracking~\cite{KIM:2014:BMVC}, optical flow estimation~\cite{Bardow:CVPR:2016}, and pose estimation~\cite{censi2014},~\cite{mueggler2014},~\cite{Kim:ECCV:2016}. Very high dynamic range of DVS makes it suitable for real-world applications.

\begin{figure}[t!]
\centering
\includegraphics[width =0.55\linewidth]{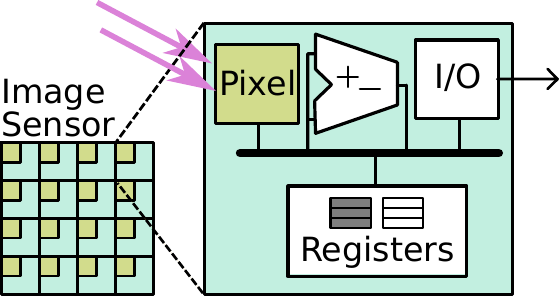}
\caption{Focal-plane Sensor-Processor Arrays (FPSPs) are parallel processing systems, where each pixel has a processing element.}
\label{fig:cpa}
\end{figure}

Cellular vision chips, such as the ACE400~\cite{dominguez1997}, ACE16K~\cite{linan2002}, MIPA4K~\cite{poikonen2009}, and Focal-plane Sensor-Processor Arrays (FPSPs)~\cite{dudek2005general},~\cite{dudek2005impl},~\cite{carey2013}, integrate sensing and processing in the focal plane. FPSPs are massively parallel processing systems on a single chip. By eliminating the need for data transmission, not only the effective frame rate is increased, but also the power consumption is reduced significantly. 
The individual processing elements are small general purpose analogue processors with a reduced instruction set and memory. Fig.~\ref{fig:cpa} shows a concept diagram of FPSP, where each pixel not only has a light-sensitive sensor, but also has a simple processing element. The main advantages of FPSPs are the high effective frame rates at lower clock frequencies which in turn reduces power consumption compared to conventional sensing and processing systems~\cite{Zhang_IEEE_2011}. However with the limited instruction sets and local memory~\cite{carey2013}, developing new applications for FPSPs, such as image filtering or camera tracking, is a challenging problem.


In the past, several interesting works have been presented using FPSPs, including high-dynamic range imaging~\cite{Martel:2016:ISCAS}. New directions are being followed to explore the performance of FPSPs in real-world robotic and virtual reality applications. These directions include 
1) 4-DOF camera tracking~\cite{Bose2017ICCV}, 
and 
2) automatic filter kernel code generation 
as well as Viola-Jones~\cite{Viola_ICCV_2001} 
face detection~\cite{Debrunner_ICRA_2018_auto}. The key concept behind these works with FPSP is the fact that FPSP is able to report sum of intensity values of all (or a selection of) pixels in just one clock cycle. This ability allows us to develop kernel code generation and also develop/verify motion hypotheses for visual odometry and camera tracking applications. The results of these works demonstrate that FPSPs not only consume much less power compared to conventional cameras, but also can be operated at very high frame rates, such as 10,000 FPS. 

\begin{figure}
\centering
\includegraphics[width=\hsize]{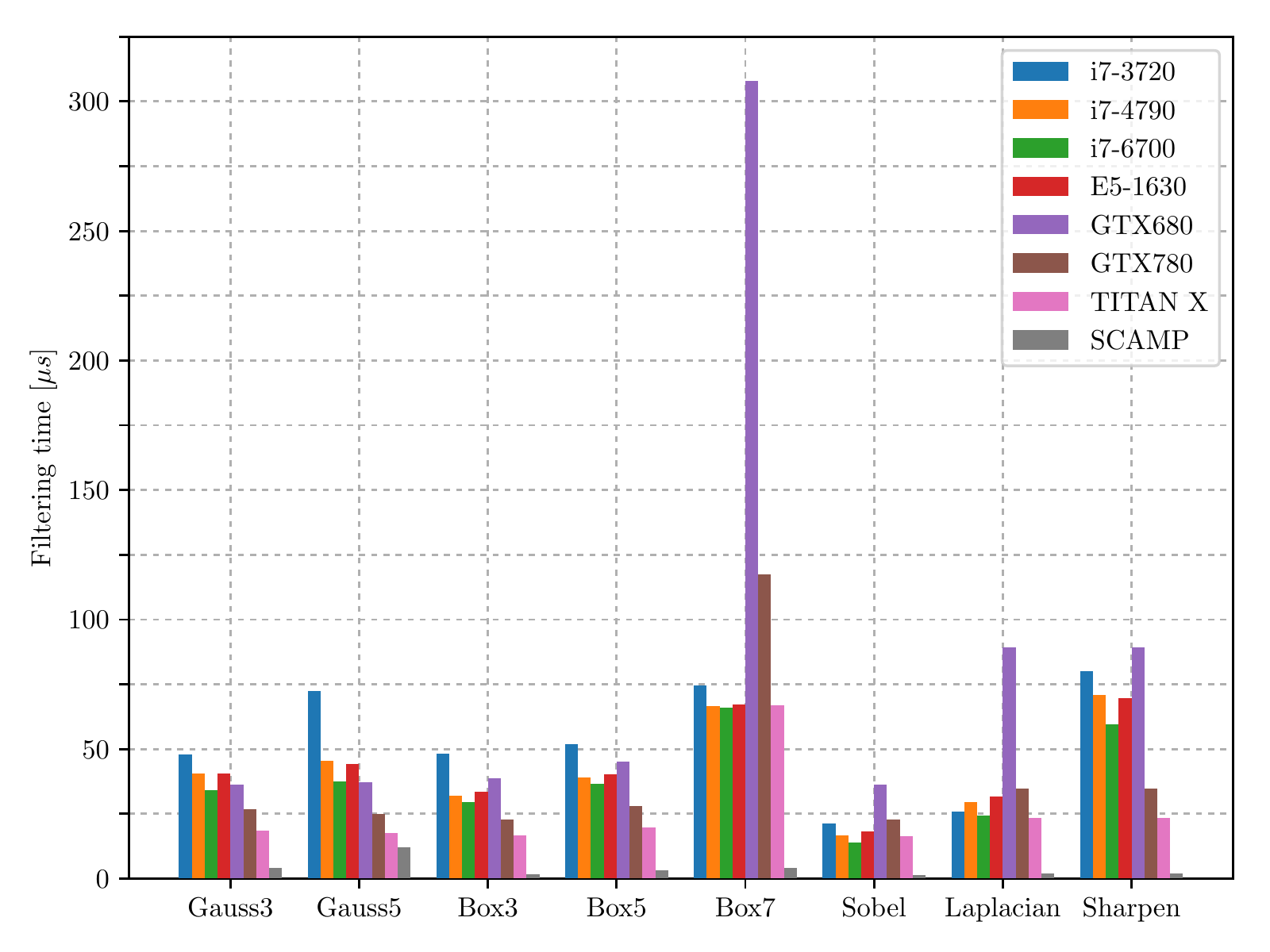}
\caption{Time for a single filter application of several well-known filters on CPU, GPU, and SCAMP FPSP hardware. The FPSP code was generated by the method explained in~\cite{Debrunner_ICRA_2018_auto}, the CPU and GPU code are  based on OpenCV 3.3.0.}
\label{fig:cpacpu_speed}
\end{figure}

Fig.~\ref{fig:cpacpu_speed} demonstrates execution times for common convolution filters on various CPUs and GPUs compared with an implementation of FPSP, known as SCAMP~\cite{carey2013}. The code for FPSP was automatically generated as explained in~\cite{Debrunner_ICRA_2018_auto}.  The parallel nature of the FPSP allows it to perform all of the tested filter kernels, shown on x-axis, in a fraction of the time needed by the other devices, shown on y-axis. This is a direct consequence of having a dedicated processing element available for every pixel, building up the filter on the whole image at the same time. 
As for the other devices, we see that for dense kernels (\textit{Gauss, Box}), GPUs usually perform better than CPUs, whereas for sparse kernels (\textit{Sobel, Laplacian, Sharpen}), CPUs seem to have an advantage. 
An outlier case being the $7\times7$ box filter, at which only the most powerful graphics card manages to get a result comparable to the CPUs. It is assumed that the CPU implementation follows a more suitable algorithm than the GPU implementation, even though both implementations are based on their vendors performance libraries (\textit{Intel IPP, nVidia NPP}). Another reason could be the fact, that the \textit{GTX680} and \textit{GTX780} are based on a  hardware architecture that is less suitable for this type of filter than the \textit{TITAN X}'s architecture. 
While Fig.~\ref{fig:cpacpu_speed} shows that there is a significant reduction in \emph{execution time}, the SCAMP chip consumes only $1.23 W$ under full load. Compared to the experimented CPU and GPU systems, this at least 20 times less \emph{power}.
Clearly a more specialised image processing pipeline architecture could be more energy-efficient than these fully programmable architectures. There is scope for further research to map the space of alternative designs, including specialised heterogeneous multicore vision processing accelerators such as the Myriad-2 Vision Processing Unit~\cite{MovidiusMyriad2}.




\section{Software Optimisations, Compilers and Runtimes} \label{sec:software}

\begin{figure}[t!]
\centering
\includegraphics[width = 0.85\linewidth]{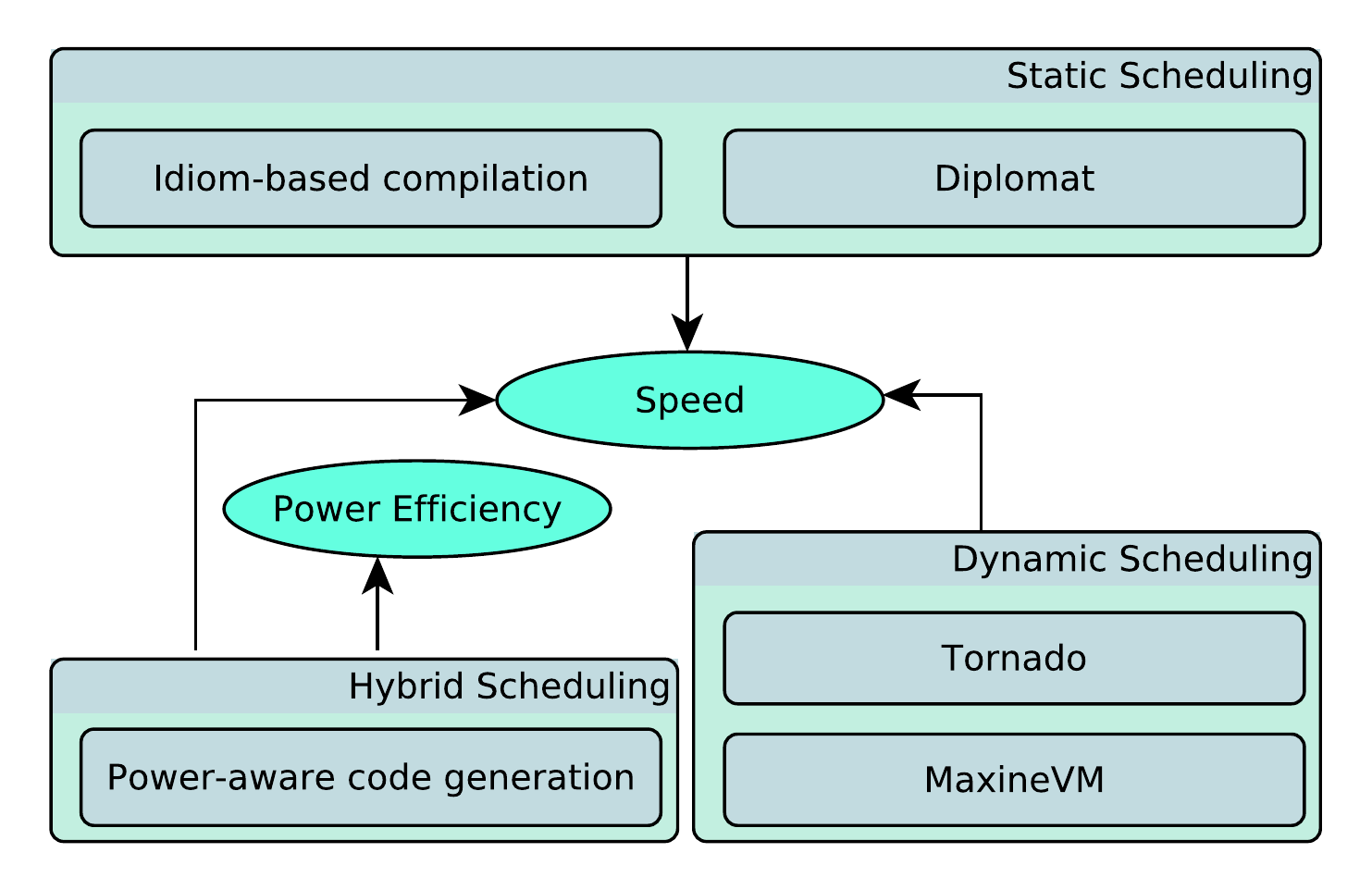}
\caption{Static, dynamic, and hybrid scheduling are the software optimisation methods presented for power efficiency and speed improvement.}
\label{fig:ManSW}
\end{figure}

In this section, we investigate how software optimisations, that are mainly implemented as a collection of compiler and runtime techniques, can be used to deliver potential improvements in power consumption and speed trade-offs.
The optimisations must determine how to efficiently map and schedule program parallelism onto multi-core, heterogeneous processor architectures.
This section presents the novel {\em static}, {\em dynamic}, and {\em hybrid} approaches used to specialise computer vision applications for execution on energy efficient runtimes and hardware (Fig.~\ref{fig:ManSW}). 

\subsection{Static Scheduling and Code Transformation}\label{sec:staticscheduling}

In this section, we focus on static techniques applied when building an optimised executable.
Static schedulers and optimisers can only rely on performance models of underlying architectures or code to optimise, which limit opportunities.
However they do not require additional code to execute, which reduces runtime overhead. 
We first introduce in~\ref{sssec:idiom} an idiom-based heterogeneous compilation methodology which given the source code of a program, can automatically identify and transform portions of code in order to be accelerated using many-core CPUs or GPUs.
Then in 
\ref{sssec:diplomat}, we propose a different methodology used to determine which resources should be used to execute those portions of code. 
This methodology takes a specialised direction, where applications need to be expressed using a particular model in order to be scheduled. 

\subsubsection{Idiom-based heterogeneous compilation}\label{sssec:idiom}
A wide variety of high-performance accelerators now exist, ranging from embedded 
DSPs, to GPUs, to highly specialised devices such as 
Tensor Processing Unit~\cite{TPU} and Vision Processing Unit~\cite{MovidiusMyriad2}. 
These devices have the capacity to deliver high performance and energy efficiency, 
but these improvements come at a cost: to obtain peak performance, the target 
application or kernel often needs to be rewritten or heavily modified. Although 
high-level abstractions can reduce the cost and difficulty of these modifications, 
these make it more difficult to obtain peak performance. In order to extract 
the maximum performance from a particular accelerator, an application must be 
aware of its exact hardware parameters (number of processors, memory sizes, bus 
speed, Network-on-Chip (NoC) routers, etc.), and this often requires low level programming and tuning.
Optimised numeric libraries and Domain Specific Languages (DSLs) have been 
proposed as a means 
of reconciling programmer ease and hardware performance. 
However, they still require significant legacy code modification and increase the 
number of languages programmers need to master.

Ideally, the compiler should be able to automatically take advantage of these
accelerators, by identifying opportunities for their use, and then automatically
calling into the appropriate libraries or DSLs. However, in practice, compilers
struggle to identify such opportunities due to the complex and expensive analysis required. Additionally, when such opportunities are found, they are frequently on a too
small scale 
to obtain any real benefit, with the cost of setting up the accelerator
(i.e. data movement, Remote Procedure Call (RPC) costs
, etc.) being much greater than the improvement in
execution time or power efficiency. Larger scale opportunities are difficult to 
identify due to the complexity of analysis, which often requires inter-procedural
analyses, loop invariant detection, pointer and alias analyses, etc., which are
complex to implement in the compiler and expensive to compute.
On the other hand, when humans attempt to use
these accelerators, they often lack the detailed knowledge of the compiler, and
resort to ``hunches'' or ad-hoc methods, leading to sub-optimal performance.

In~\cite{Ginsbach2018}, we develop a novel approach to automatically detect and
exploit opportunities to take advantage of accelerators and DSLs. We call these
opportunities ``idioms''. By expressing these idioms as constraint problems, we
can take advantage of constraint solving techniques (in our case a Satisfiability
Modulo Theories (SMT) solver). Our technique converts the constraint problem which
describes each idiom into an LLVM compiler pass. When running on LLVM IR (Intermediate Representation), these passes identify and report instances of each idiom. This technique is further strengthened by the use of Symbolic Execution and Static Analysis techniques, so that formally proved transformations can be automatically applied when idioms are detected.

We have described idioms for sparse and dense linear algebra, and stencils and reductions,
and written transformations from these idioms to the established cuSPARSE and clSPARSE
libraries, as well as a data-parallel, functional DSL which can be used to generate
high performance platform specific OpenCL code. We have then evaluated this technique on
the NAS, Parboil, and Rodinia sequential C/C++ benchmarks, where we detect 55 instances of our
described idioms. 
The NAS, Parboil, and Rodinia benchmarks include several key and frequently used computer vision and SLAM related tasks such as convolution filtering, particle filtering, backpropagation, k-means clustering, breadth-first search, and other fundamental computational building blocks.
In the cases where these idioms form a significant part of the sequential
execution time, we are able to transform the program to obtain performance improvements 
ranging from 1.24x to over 20x on integrated and discrete GPUs, contributing to the fast \emph{execution time} objective.

\subsubsection{Diplomat, Static mapping of multi-kernel applications on heterogeneous platforms} \label{sssec:diplomat} \completedby{Edinburgh - Bruno}

\begin{figure}[t!]
\centering
\includegraphics[width =.75 \linewidth]{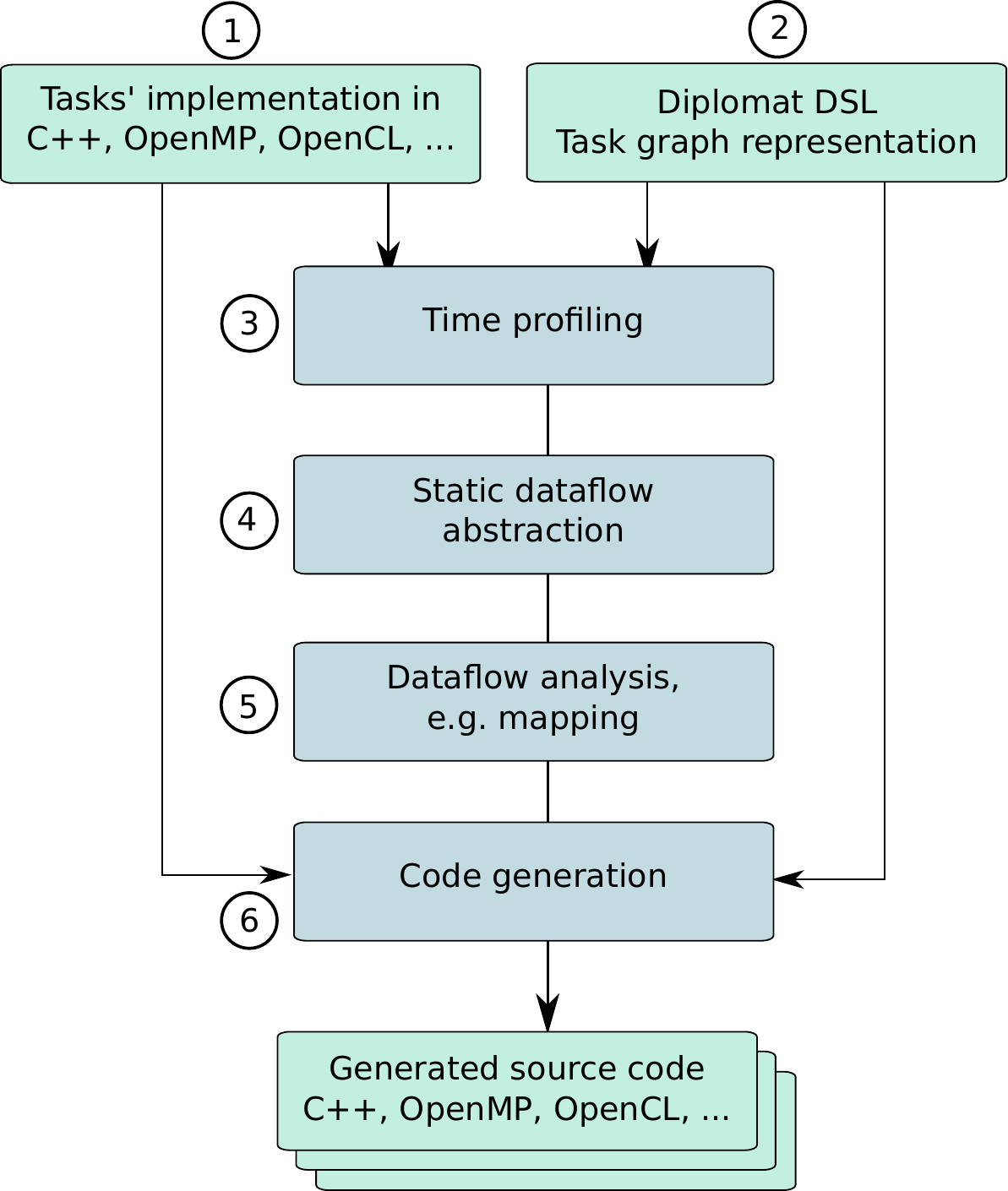}
\caption{An overview of the Diplomat framework. The user provides (1) the task implementations in various languages and (2) the dependencies between the tasks. Then in (3) Diplomat performs timing analysis on the target platform and in (4) abstracts the task-graph as a static dataflow model. Finally, a dataflow model analysis step is performed in (5), and in (6) the Diplomat compiler performs the code generation.}
\label{fig:diplomat}
\end{figure}

We propose a novel approach to heterogeneous embedded systems programmability using 
a task-graph based DSL 
called Diplomat~\cite{Bodin2016:MASCOTS16}. Diplomat is a task-graph framework that exploits the potential of static dataflow modelling and analysis to deliver performance estimation and CPU/GPU mapping. An application has to be specified once, and then the framework can automatically propose good mappings. This work aims at improving runtime as much as performance robustness.

The Diplomat front-end is embedded in the Python programming language and it allows the framework to gather fundamental information about the application: the different possible implementations of the tasks, their expected input and output data sizes, and the existing data dependencies between each of them.

At compile-time, the framework performs static analysis. In order to benefit from existing dataflow analysis techniques, the initial task-graph needs to be turned into a dataflow model.
As the dataflow graph will not be used to generate the code, a representation of the application does not need to be precise.
But it needs to model an application's behaviour close enough to obtain good performance estimations.
Diplomat performs the following steps. First, the initial task-graph is abstracted into a static dataflow formalism. This includes a timing profiling step to estimate task durations and communication delays. Then, by using static analysis techniques~\cite{Bodin2016:DAC16}, a throughput evaluation and a mapping of the application are performed.

Once a potential mapping has been selected, an executable C++ code is automatically generated. This generated implementation takes advantage of task-parallelism and data-parallelism. It can use OpenMP and OpenCL and it may apply partitioning between CPU and GPU when it is beneficial. This overview is summarised in Fig.~\ref{fig:diplomat}.

\begin{figure}[t!]
\centering
\includegraphics[width =.72 \linewidth]{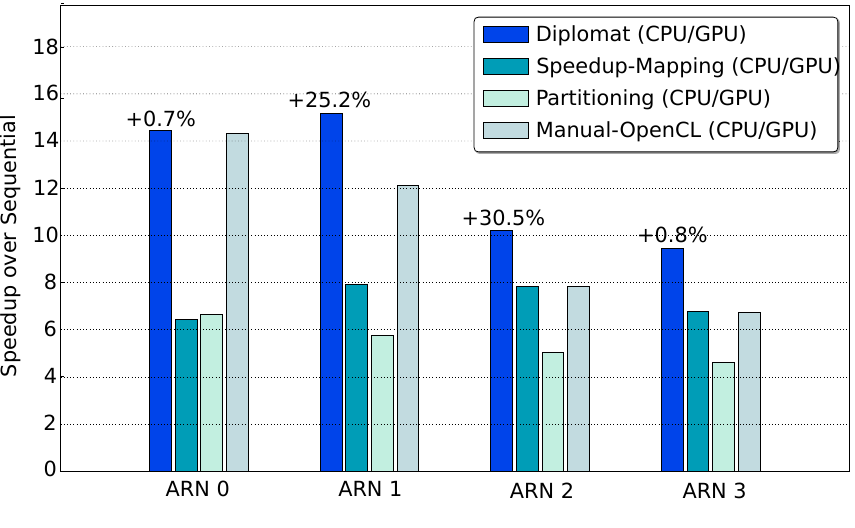}
\caption{Evaluation of the best result obtained with Diplomat for CPU and GPU configurations, and comparison with handwritten solutions (OpenMP, OpenCL) and automatic heuristics (Partitioning, Speed-up mapping) for KinectFusion on Arndale platform. The associated numbers on x-axis are different KinectFusion algorithmic parameter configuration, and the percent on top of Diplomat bars are the speedup over the manual implementation.}
\label{fig:diplomat_fig8b}
\end{figure}

We evaluate Diplomat with KinectFusion on two embedded platforms, Odroid-XU3 and Arndale, with four different configurations for algorithmic parameters, chosen manually. Fig.~\ref{fig:diplomat_fig8b} shows the results for Arndale for four different configurations, marked as ARN0...3.  
Using Diplomat, we observed a 16\% speed improvement on average and up to a 30\% improvement over the best existing hand-coded implementation. This is an improvement on \emph{runtime speed}, one of the goals outlined earlier.

\subsection{Dynamic Scheduling}\label{sec:dynamicscheduling}
Dynamic scheduling takes place while the optimised program runs with actual data. Because dynamic schedulers can monitor actual performance, they can compensate for performance skews due to data-dependant control-flow and computation that static schedulers cannot accurately capture and model. Dynamic schedulers can therefore exploit additional dynamic run-time information to enable more optimisation opportunities. However, they also require the execution of additional profiling and monitoring code, which can create performance penalties.

Tornado and MaxineVM 
runtime scheduling are research prototype systems that we are using to explore and investigate dynamic scheduling opportunities. 
Tornado is a framework (prototyped on top of Java) using dynamic scheduling for transparent exploitation of task-level parallelism on heterogeneous systems having multicore CPUs, and accelerators such as GPUs, DSPs and FPGAs. 
MaxineVM is a research Java Virtual Machine (JVM) that we are initially using to investigate dynamic heterogeneous multicore scheduling for application and JVM service threads in order to better meet the changing power and performance objectives of a system under dynamically varying battery life and application service demands. 

\subsubsection{Tornado}\label{sec:tornado}
\completedby{Manchester - Andy} Tornado is a heterogeneous programming framework that has been designed for programming systems that have a higher-degree of heterogeneity than existing GPGPU accelerated systems and where system configurations are unknown until runtime. The current Tornado prototype~\cite{maxineVEE}  
superseding 
JACC, described in~\cite{jaccPREPRINT}, can dynamically offload code to big.LITTLE cores, and GPUs with its OpenCL backend that supports the widest possible set of accelerators. Tornado can also be used to generate OpenCL code that is suitable for high-level synthesis tools in order to produce FPGA accelerators, although it is not practical to do this unless the relatively long place and route times of FPGA vendor tools can be amortised by application run-time overheads. The main benefit of Tornado is that it allows portable dynamic exploration of how heterogeneous scheduling decisions for task-parallel frameworks will lead to improvements in power-performance trade-offs without rewriting the application level code, and also where knowledge of the heterogeneous configuration of a system is delayed until runtime. 

\begin{figure}
\centering
\includegraphics[scale=0.5]{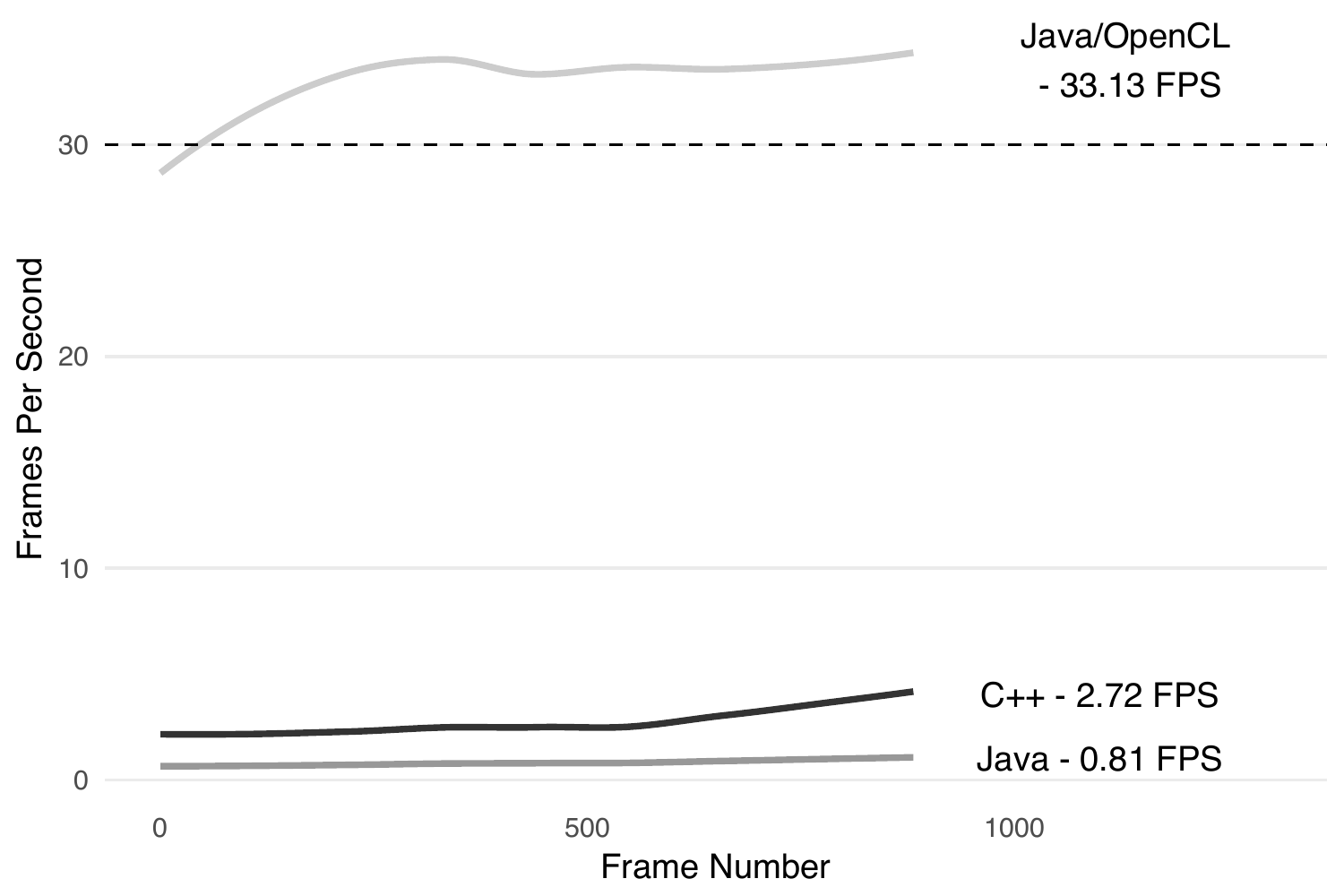}
\caption{Execution performance of KinectFusion (using FPS) over the time using Tornado (Java/OpenCL) vs. baseline Java and C++.}
\label{fig:tornado-fps}
\end{figure}

The Tornado API cleanly separates computation logic from co-ordination logic that is expressed using a task-based programming model. Currently, data parallelisation is expressed using standard Java support for annotations~\cite{maxineVEE}. Applications remain architecture-neutral, and as the current implementation of Tornado is  based on the Java managed language,  we are able to dynamically generate code for heterogeneous execution without recompilation of the Java source, and without manually generating new optimised routines for any 
accelerators that may become available. Applications need only to be configured at runtime for execution on the available hardware.  Tornado currently uses an OpenCL driver for maximum device coverage: this  includes mature support for: multi-core CPUs and GPGPU, and maturing support for Xeon Phi coprocessor/accelerators. The current dynamic compiler technology of Tornado is built upon JVMCI and GRAAL APIs for Java 8 and above. The sequential Java and C++ versions of KinectFusion in SLAMBench both perform at under 3 FPS with the C++ version being 3.4x faster than Java. This improvement of \emph{runtime speed} is shown in Fig.~\ref{fig:tornado-fps}. By accelerating
KinectFusion through GPGPU execution using Tornado, we manage to achieve a constant
rate of over 30 FPS (33.13 FPS) across all frames (882)
from the ICL-NUIM dataset with room 2 configuration~\cite{2014Handa}. To achieve 30 FPS, all kernels have been accelerated
by up to 821.20x with an average of 47.84x across the whole
application~\cite{maxineVEE, morevmstornado}. Tornado is an attractive framework for the development of portable computer vision applications as its dynamic JIT compilation for traditional CPU cores and OpenCL compute devices such as GPUs  enables real-time performance constraints to be met whilst eliminating the need to rewrite and optimise code for different GPU devices.

\subsubsection{MaxineVM} \label{sec:maxinevm}
\completedby{Manchester - Andy} 
The main contribution of MaxineVM is to provide a research infrastructure for managed runtime systems that can execute on top of modern Instruction Set Architectures (ISA)s supplied by both Intel and ARM. 
This is especially relevant because ARM is the dominant ISA in mobile and embedded platforms.  
MaxineVM has been released as open-source software~\cite{beehiveMAXSIMgithub}.

Heterogeneous multicore systems comprised of CPUs having the same ISA but different power/performance design point characteristics create a significant challenge for virtual machines that are typically agnostic to CPU core heterogeneity when undertaking thread-scheduling decisions.   Further, heterogeneous CPU core clusters, are typically attached to NUMA-like memory system designs, consequently thread scheduling policies need to be adjusted to make appropriate decisions that do not adversely affect the performance and power consumption of managed applications.  

In MaxineVM, we are using the Java managed runtime environment to optimise thread scheduling for heterogeneous architectures. Consequently, we have chosen to use and extend the Oracle Labs research project software for MaxineVM~\cite{maxineTACO} that provided a state-of-the-art research VM for x86-64. We have developed a robust port of MaxineVM to ARMv7~\cite{maxineVEE, morevmsmaxine} (an AArch64 port is also in progress) ISA processors that can run important Java and SLAM benchmarks, including a Java version of KinectFusion. 
MaxineVM has been designed for maximum flexibility, this sacrifices some performance, but it is trivially possible to replace the public implementation of an interface or {\em scheme}, such as for monitor or garbage collection with simple command line switches to the command that generates a MaxineVM 
executable image.

\subsection{Hybrid Scheduling}\label{sec:hybridscheduling}
Hybrid scheduling considers dynamic techniques which takes advantage of static and dynamic data. 
A schedule can be statically optimised for a target architecture and application (i.e. using machine learning), and a dynamic scheduler can further adjust this schedule to optimise further actual code executions. 
Since it can rely on a statically optimised schedule, the dynamic scheduler can save a significant amount of work and therefore lower its negative impact on performance.


\subsubsection{Power-aware Code Generation} 

\completedby{Edinburgh - Bruno}

Power is an important constraint in modern multi-core processor design. 
We have shown that power across heterogeneous cores varies considerably~\cite{Chandramohan2014:LCTS14}.
This work develops a compiler-based approach to runtime power management for heterogeneous cores.
Given an externally determined power budget, it generates parallel parameterised partitioned code that attempts to give the best performance within that power budget.
It uses the compiler infrastructure developed in~\cite{Chandramohan2014:CASES14}. 
The hybrid scheduling has been tested on standard benchmarks such as DSPstone, UTSDP, and Polybench. These benchmarks provide an in-depth comparison with other methods and include key building blocks of many SLAM and computer vision tasks such as matrix multiplication, edge detection, and image histogram.
We applied this technique to embedded parallel OpenMP benchmarks on the TI OMAP4 platform for a range of power budgets. 
On average we obtain a 1.37x speed-up over dynamic voltage and frequency scaling (DVFS). 
For low power budgets, we see a 2x speed-up improvement. 
SLAM systems, and vision applications in general, are composed of different phases. An adaptive power budget for every phases positively impact frame rate and power consumption.

\section{Hardware and Simulation} \label{sec:hardware}
The designers of heterogeneous Multiprocessor System-on-Chip (MPSoC) are faced with an enormous task when attempting to design a system that is co-optimised to deliver power-performance efficiency under a wide range of dynamic operating conditions concerning the available power stored in a battery, and the current application performance demands. 
In this paper, a variety of simulation tools and technologies have been presented to assist designers in their evaluations of how performance, energy, and power consumption trade-offs are affected by computer vision algorithm parameters and computational characteristics of specific implementations on different heterogeneous processors and accelerators. Tools have been developed that focus on the evaluation of native and managed runtime systems, that execute on ARM and x86-64 processor instruction set architectures in conjunction with GPU and custom accelerator intellectual property.

\begin{figure}[t]
\centering
\includegraphics[width = .85\linewidth]{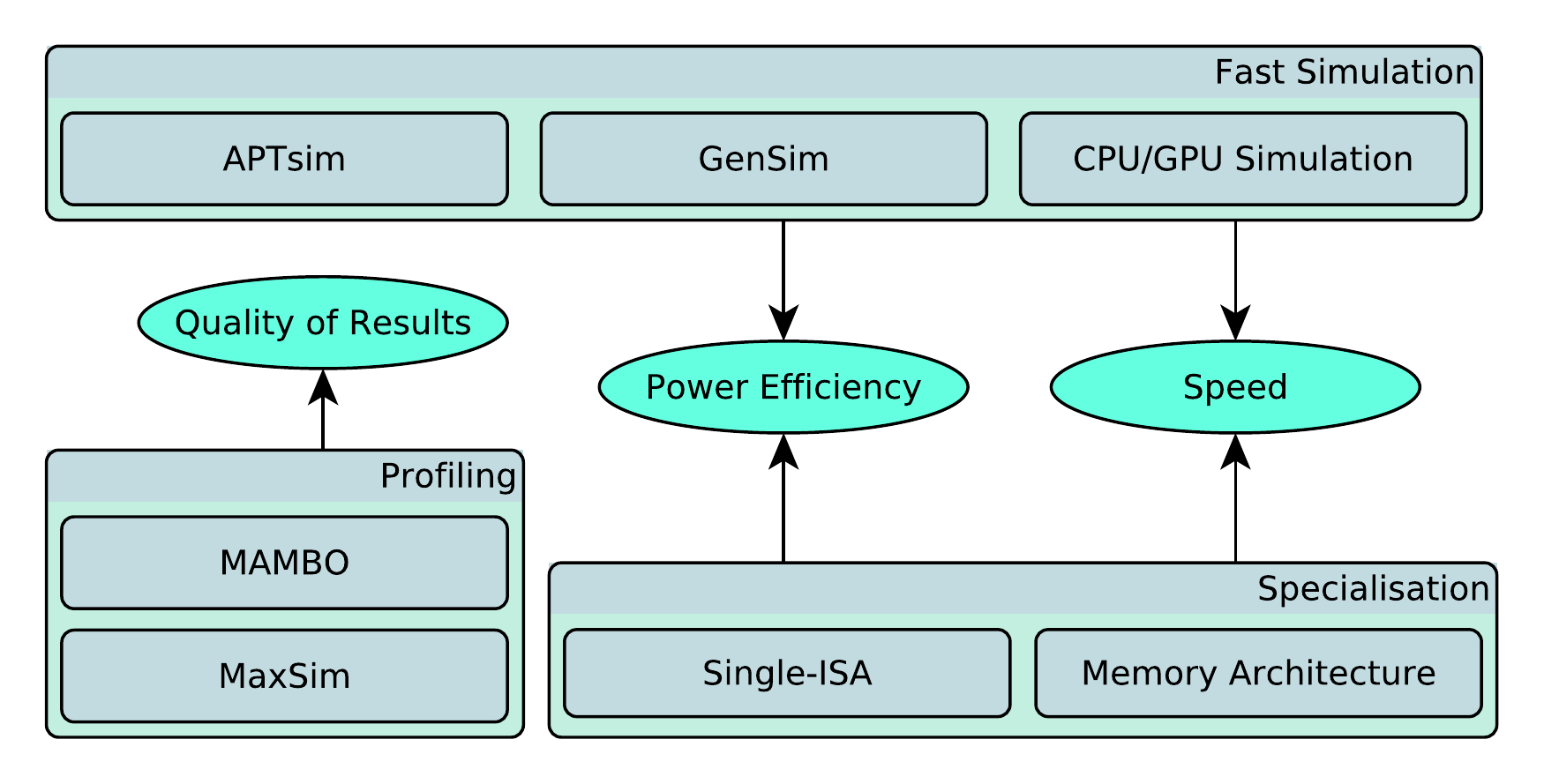}
\caption{Hardware development tasks are simulation, profiling, and specialisation tools; each with its own goals. With these three task, it is possible to develop customised hardware for computer vision applications.}
\label{fig:ManHW}
\end{figure}

The contributions of this section have been organised under three main topics: \emph{simulation}, \emph{profiling}, and \emph{specialisation}. Under each topic, several novel tools and methods are presented. The main objective in developing these tools and methods is to reduce development complexity and increase reproducibility for system analysis. Fig.~\ref{fig:ManHW} presents a graph where all simulation, profiling, and specialisation tools are summarised.

\subsection{Fast Simulation}\label{sec:simulation}
Simulators have become an essential tool for hardware design. They allow designers to prototype different systems before committing to a silicon design, and save enormous amounts of money and time. They allow embedded systems engineers to develop the driver and compiler stack, before the system is available, and be able to verify their results. Even after releasing the hardware, software engineers can make use of simulators to prototype their programs in a virtual environment, without the latency of flashing the software onto the hardware, or even without access to the hardware.

These different use cases require very different simulation technologies. Prototyping hardware typically requires `detailed' performance modelling simulation to be performed, which comes with a significant slowdown compared to real hardware. On the other hand, software development often does not require such detailed simulation, and so faster `functional' simulators can be used. This has led to the development of multiple simulation systems within this work, with 
the GenSim system being used for `functional' simulation and APTsim being used for more detailed simulation.


In this section, three novel system simulation works are presented. These works are: GenSim, CPU/GPU simulation, and APTsim.

\subsubsection{The GenSim Architecture Description Language}\completedby{Edinburgh - Harry}
              
Modern CPU architectures often have a large number of extensions and versions. At the same time, simulation technologies have improved, making simulators both faster and more accurate. However, this has made the creation of a simulator for a modern architecture much more complex. Architecture Description Languages (ADLs) seek to solve this problem by decoupling the details of the simulated architecture from the tool used to simulate it. 

\begin{figure}[t!]
\centering
\includegraphics[width=0.85\columnwidth]{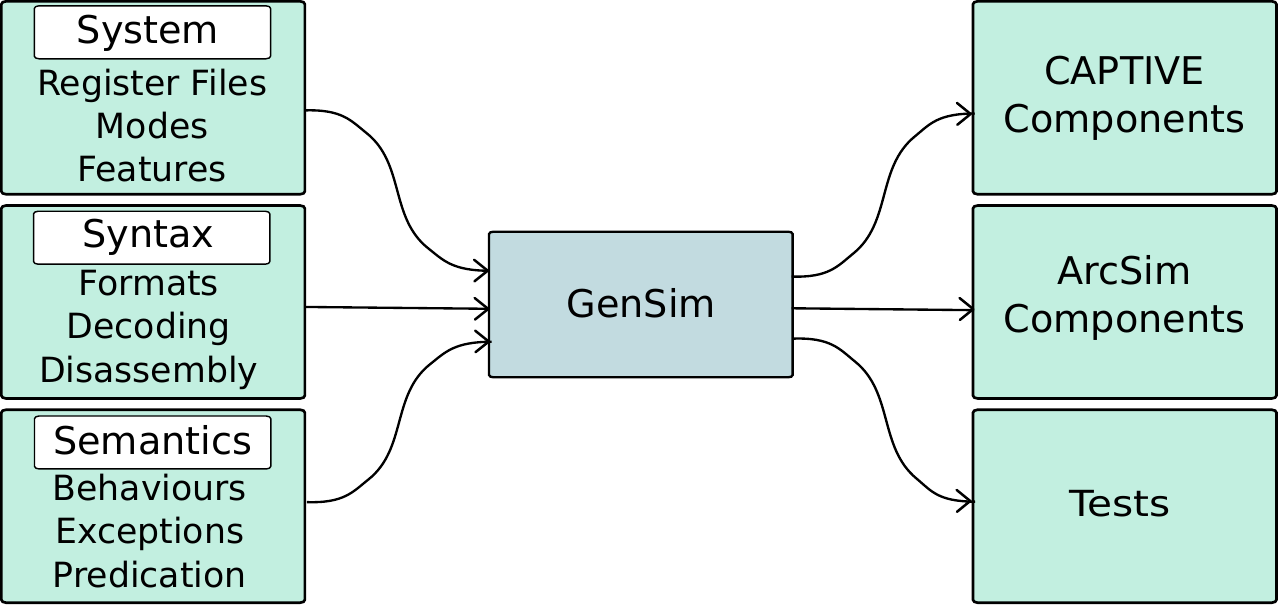}
\caption{Diagram showing the general flow of the GenSim ADL toolchain}
\label{fig:gensim}
\end{figure}

We have developed the GenSim simulation infrastructure, which includes an ADL toolchain (see Fig.~\ref{fig:gensim}). This ADL is designed to enable the rapid development of fast functional simulation tools~\cite{spink2014efficient}, and the prototyping of architectural extensions (and potentially full instruction set architectures). This infrastructure is used in 
the CPU/GPU simulation work (Section~\ref{sec:cpugpu}). The GenSim infrastructure is described in a number of publications~\cite{Wagstaff:2013:EPE:2463209.2488760, wagstaff2014automated,spink2015efficient}. 
GenSim is available under a permissive open-source license, and is available at~\cite{gensimwebsite}.

\subsubsection{Full-system simulation for CPU/GPU}  \completedby{Edinburgh - Harry}
\label{sec:cpugpu}


        
Graphics processing units 
are highly-specialized processors that were originally designed to process large graphics workloads effectively, however they have been influential in many industries, including in executing computer vision tasks. Simulators for parallel architectures, including GPUs, have not reached the same level of maturity as simulators for CPUs, both due to the secrecy of leading GPU vendors, and the problems arising from mapping parallel onto scalar architectures, or onto different parallel architectures. 

At the moment, GPU simulators that have been presented in literature have limitations, resulting from lack of verification, poor accuracy, poor speeds, and limited observability due to incomplete modelling of certain hardware features. As they don't accurately model the full native software stack, they are unable to execute realistic GPU workloads, which rely on extensive interaction with user and system runtime libraries. 

In this work, we propose a full-system methodology for GPU simulation, where rather than simulating the GPU as an independent unit, we simulate it as a component of a larger system, comprising a CPU simulator with supporting devices, operating system, and a native, unmodified driver stack. This faithful modelling results in a simulation platform indistinguishable from real hardware. 

We have been focusing our efforts on simulation of the ARM Mali GPU, and have built a substantial amount of surrounding infrastructure. We have seen promising results in simulation of compute applications, most notably SLAMBench.

The work directly contributed to full system simulation, by implementing the ARMv7 MMU, ARMv7 and Thumb-2 Instruction Sets, and a number of devices needed to communicate with the GPU. To connect the GPU model realistically, we have implemented an ARM CPU GPU interface containing an ARM Device on the CPU side~\cite{Kaszyk2018DAC}.

The implementation of the Mali GPU simulator comprises:
\begin{itemize}
\item An implementation of the Job Manager, a hardware resource for controlling jobs on the GPU side,
\item The Shader Core Infrastructure, which allows for retrieving important context, needed to execute shader programs efficiently,
\item The Shader Program Decoder, which allows us to interpret Mali Shader binary programs,
\item The Shader Program Execution Engine, which allows us to simulate the behaviour of Mali programs.
\end{itemize}

Future plans for simulation include extending the infrastructure to support real time graphics simulation, increasing GPU Simulation performance using Dynamic Binary Translation (DBT)~\cite{spink2014efficient,spink2015efficient,spink2016efficient} techniques, and extending the Mali Model to support performance modelling. We have also continued to investigate new techniques for full-system dynamic binary translation (such as exploiting hardware features on the host to further accelerate simulation performance), as well as new methodologies for accelerating the implementation and verification of full system instruction set simulators. Fast full system simulation presents a large number of unique challenges and difficulties and in addressing and overcoming these difficulties, we expect to be able to produce a significant body of novel research. Taken as a whole, these tools will directly allow us to explore next-generation many-core applications, and design hardware that is characterised by high performance and low power.
                
\subsubsection{APTSim - simulation and prototyping platform} 
\completedby{Manchester - John}

\begin{figure}[t!]
\centering
\includegraphics[width =.8\linewidth]{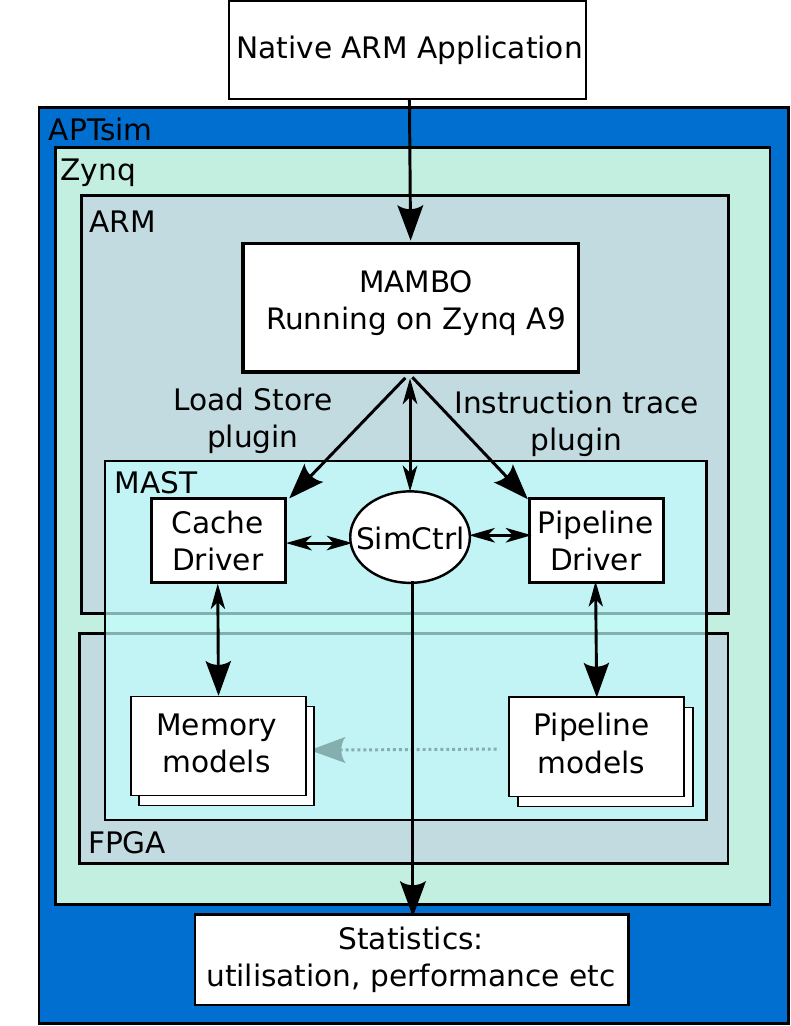}
\caption{APTSim an FPGA accelerated simulation and prototyping platform, currently implemented on Zynq SoC.}
\label{fig:APTSim}
\end{figure}

APTSim (Fig.~\ref{fig:APTSim}) is intended as a fast simulator allowing rapid  simulation of microprocessor architectures and microarchitectures as well as the prototyping of accelerators. The system runs on a platform consisting of a processor, for functional simulation, and an FPGA for implementing architecture timing models and prototypes. Currently the Xilinx Zynq family is used as the host platform.  APTSim performs dynamic binary instrumentation using MAMBO, see Section~\ref{mambo}, to dynamically instrument a running executable along with the MAST co-design library, described below. Custom MAMBO plugins allow specific instructions, such as load/store or PC changing events to be sent to MAST hardware models, such as memory systems or processor pipeline. From a simulation perspective the hardware models are for timing and gathering statistics and do not perform functional simulation, which is carried out on the host processor as native execution; so for example if we send a request to a cache system the model will tell us at which memory level the result is present in and a response time, while the actual data will be returned from the processor's own memory. This separation allows for smaller, less complicated, hardware models to gather statistics whilst the processor executes the benchmark natively and the MAMBO plugins capture the necessary events with low overhead.

The MAST library provides a framework for easily integrating many hardware IP blocks, implemented on FPGA,  with a linux based application running on a host processor. MAST consists of two principal parts: a software component and a hardware library. The software component allows the discovery and management of hardware IP blocks and the management of memory used by the hardware blocks; critically this allows new hardware blocks to be configured and used at runtime using only user space software.  The  hardware library, written in Bluespec, contains parametrised IP blocks including architecture models such as cache systems or pipeline models and accelerator modules for computer vision, such as filters or feature detectors. The hardware models can either be masters or slaves, from a memory perspective. As masters, models can directly access processor memory leaving the processor to execute code whilst the hardware is analysing the execution of the last code block.

APTSim also allows us to evaluate prototype hardware, for example we evaluated multiple branch predictors by implementing them in Bluespec and using a MAST compliant interface. This allows us to execute our benchmark code once on the CPU and offload to multiple candidate implementations to rapidly explore the design space. 


In~\cite{Mawer2017:ICCM} we show that on the Xilinx Zynq 7000 FPGA board coupled with a relatively slow 666MHz ARM9 processor, the slowdown of APTsim is 400x in comparison to native execution on the same processor. While a relatively important slowdown over native execution is unavoidable to implement a fine performance monitoring, slowdown on APTsim is about half of GEM5 running at 3.2GHz on an Intel Xeon E3 to simulate the same ARM system. Note that, contrary to APTsim, GEM5 on Xeon does not take profit of any FPGA acceleration. This shows the interest of APTsim to take profit of FPGA acceleration to implement a fast Register Transfer Level (RTL) simulation and monitor its performance, while hiding the complexity of FPGA programming from the user.

\subsection{Profiling} \label{sec:profiling}


Profiling is the process of analysing the runtime behaviour of a program in order to perform some measurements about the performance of the program. For example, to determine which parts of the program take the most time to execute. This information can then be used to improve software (for example, by using a more optimised implementation of frequently executed functions) or to improve hardware (by including hardware structures or instructions which provide better performance for frequently executed functions). Profiling of native applications is typically performed via dynamic binary instrumentation. However, when a managed runtime environment is used, the runtime environment can often perform the necessary instrumentation. In this section, we explore both of these possibilities, with MAMBO being used for native profiling, and MaxSim being used for the profiling of Java applications.


\subsubsection{MAMBO: instruction level profiling} \label{mambo}\completedby{Manchester - Cosmin}

Dynamic Binary Instrumentation (DBI) is a technique for instrumenting applications transparently while they are executed, working at the level of machine code. As the ARM architecture expands beyond its traditional embedded domain, there is a growing interest in DBI systems for the general-purpose multicore processors that are part of the ARM family. DBI systems introduce a performance overhead and reducing it is an active area of research; however, most efforts have focused on the x86 architecture.

MAMBO is a low overhead DBI framework for 32-bit (AArch32) and 64-bit ARM (AArch64)~\cite{mambo_p1}. MAMBO is open-source~\cite{gorgovan2016mambogithub}. MAMBO provides an event-driven plugin API for the implementation of instrumentation tools with minimal complexity. The API allows the enumeration, analysis and instrumentation of the application code ahead of execution, as well as tracking and control of events such as system calls. Furthermore, the MAMBO API provides a number of high level facilities for developing portable instrumentation, i.e. plugins which can execute efficiently both on AArch32 and AArch64, while being implemented using mostly high level architecture-agnostic code.

MAMBO incorporates a number of novel optimisations, specifically designed for the ARM architecture, which allow it to minimise its performance overhead. The geometric mean runtime overhead of MAMBO running SPEC CPU2006 with no instrumentation is as low as 12\% (on an APM X-C1 system), compared DynamoRIO~\cite{bruening2004efficient}, a state of the art DBI system, which has an overhead of 34\% under the same test conditions.

\begin{figure}[t!]
\centering
\includegraphics[width =.8 \linewidth]{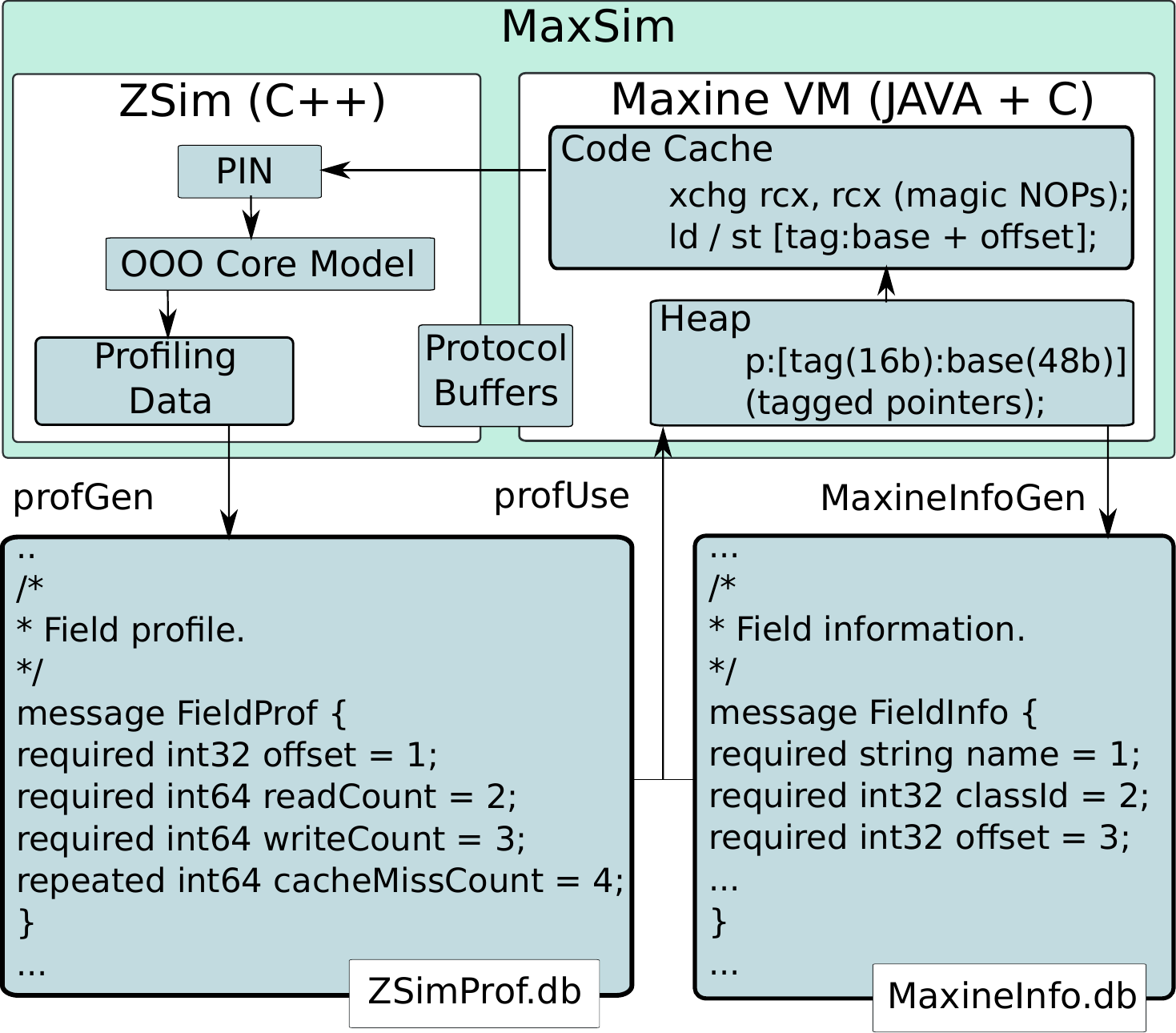}
\caption{MaxSim overview of Zsim and MaxineVM based profiling}
\label{fig:maxsim}
\end{figure}

\subsubsection{MaxSim: profiling and prototyping hardware-software co-design for managed runtime systems}  \label{sec:maxsim}
\completedby{Manchester - Andy} 

Managed applications, written in programming languages such as Java, C\# and others, represent a significant share of workloads in the mobile, desktop, and server domains. Microarchitectural timing simulation of such workloads is useful for characterisation and performance analysis, of both hardware and software, as well as for research and development of novel hardware extensions. MaxSim~\cite{maxsimISPASS} (see Fig.~\ref{fig:maxsim}), is a simulation platform based on the MaxineVM~\cite{maxineTACO} (explained in Section~\ref{sec:maxinevm}), the ZSim~\cite{zsimISCA} simulator, and the McPAT~\cite{mcpatMICRO} modelling framework. MaxSim can perform fast and accurate simulation of managed runtime workloads running on top of Maxine VM~\cite{beehiveMAXSIMgithub}. MaxSim's capabilities include: 1) low-intrusive microarchitectural profiling via pointer tagging on x86-64 platforms, 2) modelling of hardware extensions related, but not limited to, tagged pointers, and 3) modelling of complex software changes via address-space morphing.
        
\begin{figure}[t!]
\centering
\includegraphics[width =.6 \linewidth]{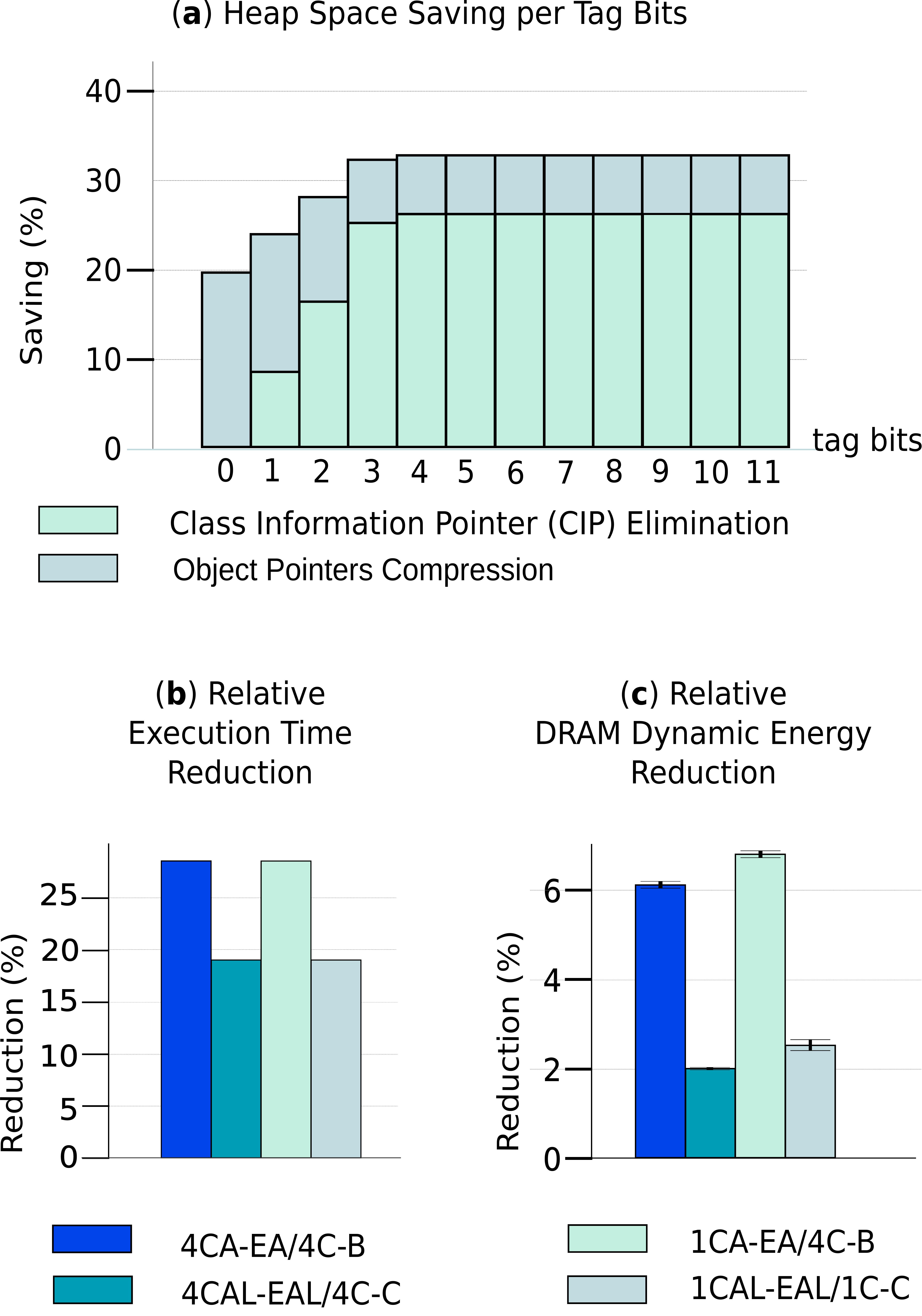}
\caption{Performance of MaxSim on KinectFusion. 
({\bf a}) heap space saving using tagged pointers,
({\bf b}) relative reduction in execution time, and
({\bf c}) relative reduction in DRAM dynamic energy.
}
\label{fig:maxsimKF}
\end{figure}
        
Low-intrusive microarchitectural profiling is achieved by utilising tagged pointers to collect type- and allocation-site related hardware events. Furthermore, MaxSim allows, through a novel technique called address space morphing, the easy modelling of complex object layout transformations. Finally, through the co- designed capabilities of MaxSim, novel hardware extensions can be implemented and evaluated.
We showcase MaxSim's capabilities by simulating the whole set of the DaCapo-9.12-bach benchmarks in less than a day while performing an up-to-date microarchitectural power and performance characterisation~\cite{maxsimISPASS}. Furthermore, we demonstrate a hardware/software co-designed optimisation that performs dynamic load elimination for array length retrieval achieving up to 14\% L1 data cache loads reduction and up to 4\% dynamic energy reduction. 
In~\cite{maxsimIEEETOC} we present results for MaxineVM with MaxSim.
We use SLAMBench to experiment with KinectFusion on a 4-core Nehalem system, using 1 and 4 cores (denoted by \emph{1C} and \emph{4C}, respectively). We use MaxSim's extensions for the Address Generation Unit (AGU) (denoted by 1CA and 4CA) and Load-Store Unit (LSU) extension (shown by 1CAL and 4CAL). Fig.~\ref{fig:maxsimKF}-a shows heap savings of more than 30\% on SLAMBench thanks to CIP (Class Information Pointer) elimination.
Fig.~\ref{fig:maxsimKF}-b demonstrates the relative reduction in execution time, using the proposed framework. On this figure, \emph{EA} refers to a machine configuration with CIP elimination with 16 bits CID (Class Information) and \emph{EAL} refers to a variant with CIP elimination, 4 bits CID, and AGU and LSU extensions. \emph{B} stands for the standard baseline MaxSim virtual machine and \emph{C} is \emph{B} with object compression. Fig~\ref{fig:maxsimKF}-b shows up to 6\% execution time performance benefits of CIP elimination over MaxSim with none of our extension, whether its uses 4 cores (\emph{4CA-EA/4CA-B}) or 1 core (\emph{1CA-EA/1C-B})
Finally, Fig.~\ref{fig:maxsimKF}-c shows the relative reduction in DRAM dynamic energy for the cases mentioned above. As the graph shows, there is an 18\% to 28\% reduction in DRAM dynamic energy. These reductions contribute to the objective of having improved \emph{quality of the results}.
MaxSim is open-source~\cite{beehiveMAXSIMgithub}.

\subsection{Specialisation} \label{sec:specialisation}




Recent developments in computer vision and machine learning have challenged hardware and circuit designers to design faster and more efficient systems for these tasks~\cite{Sze_2017_IEEE}. Tensor Processing Unit (TPU) from Google~\cite{TPU}, Vision Processing Unit (VPU) from Intel Movidius~~\cite{MovidiusMyriad2}, 
and Intelligent Processing Unit (IPU) from Graphcore~\cite{Graphcore}, are such devices with major re-engineerings in hardware design, resulting in outstanding performance. While the development of custom hardware can be appealing due to the possible significant benefits, it can lead to extremely high design, development, and verification costs, and a very long time to market. One method of avoiding these costs while still obtaining many of the benefits of custom hardware is to specialise existing hardware. We have explored several possible paths to specialisation, including specialised memory architectures for GPGPU computations (which are frequently used in computer vision algorithm implementations), the use of single-ISA heterogeneity (as seen in ARM's big.LITTLE platforms), and the potential for power and area savings by replacing hardware structures with software.

\subsubsection{Memory Architectures for GPGPU Computation} \completedby{Edinburgh - Bruno}
Current GPUs are no longer perceived as accelerators solely for graphic workloads, and now cater to a much broader spectrum of applications. In a short time, GPUs have proven to be of substantive significance in the world of general-purpose computing, playing a pivotal role in Scientific and High Performance Computing (HPC). The rise of general-purpose computing on GPUs has contributed to the introduction of on-chip cache hierarchies in those systems. Additionally, in SLAM algorithms, reusing previously processed data frequently occurs such as in bundle adjustment, loop detection, and loop closure. It has been shown that efficient memory use can improve the runtime speed of the algorithm. For instance, the Distributed Particle (DP) filter optimises memory requirements using an efficient data structure for maintaining the map~\cite{Eliazar:2003:DFR}.

We have carried out a workload characterisation of GPU architectures on general-purpose workloads, to assess the efficiency of their memory hierarchies~\cite{Dublish2016:IIWC16} and proposed a novel cache optimisation to resolve some of the memory performance bottlenecks in GPGPU systems~\cite{Dublish2016:TACO16}.

\begin{figure}[t!]
\centering
\includegraphics[width =.95 \linewidth]{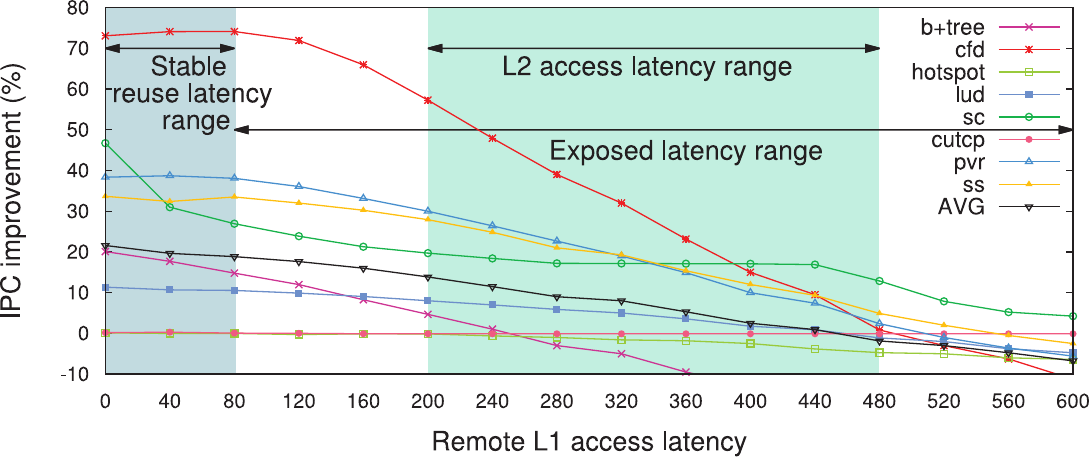}
\caption{Speed-up of Instructions Per Cycle (IPC) with varying remote L1 access latencies.}
\label{fig:taco-dublish}
\end{figure}
In our workload characterisation study (overview on Fig.~\ref{fig:taco-dublish}) we saw that, in general, high level-1 (L1) data cache miss rates place high demands on the available level-2 (L2) bandwidth that is shared by the large number of cores in typical GPUs. In particular, Fig.~\ref{fig:taco-dublish} represents bandwidth as the number of Instruction Per Cycle (IPC).  Furthermore, the high demand for L2 bandwidth leads to extensive congestion in the L2 access path, and in turn this leads to high memory latencies. Although GPUs are heavily multi-threaded, in memory intensive applications the memory latency becomes exposed due to a shortage of active compute threads, reducing the ability of the multi-threaded GPU to hide memory latency (Exposed latency range on Fig.~\ref{fig:taco-dublish}).
Our study also quantified congestion in the memory system, at each level of the memory hierarchy, and characterised the implications of high latencies due to congestion. We identified architectural parameters that play a pivotal role in memory system congestion, and explored the design space of architectural options to mitigate the bandwidth bottleneck. We showed that the improvement in performance achieved by mitigating the bandwidth bottleneck in the cache hierarchy can exceed the speedup obtained by simply increasing the on-chip DRAM bandwidth. We also showed that addressing the bandwidth bottleneck in isolation at specific levels can be suboptimal and can even be counter-productive. In summary, we showed that it is imperative to resolve the bandwidth bottleneck synergistically across all levels of the memory hierarchy. The second part of our work in this area aimed to reduce the pressure on the shared L2 bandwidth. One of the key factors we have observed is that there is significant replication of data among private L1 caches, presenting an opportunity to reuse data among the L1s. We have proposed a Cooperative Caching Network (CCN), which exploits reuse by connecting the L1 caches with a lightweight ring network to facilitate inter-core communication of shared data. When measured on a selection of GPGPU benchmarks, this approach delivers a performance improvement of 14.7\% for applications that exhibit reuse.

\subsubsection{Evaluation of single-ISA heterogeneity}  
\completedby{Edinburgh - Bruno}
\begin{figure}[t!]
\centering
\includegraphics[width =.7 \linewidth]{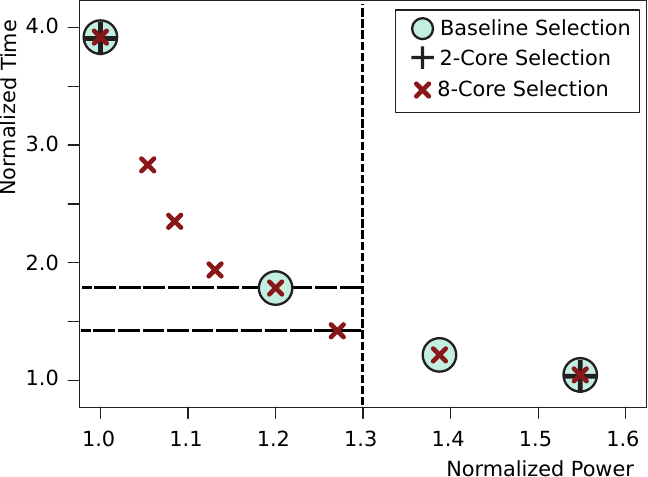}
\caption{Example of a Baseline selection, and 2- and 8-Core selections for a specific benchmark application.}
\label{fig:tomusk}
\end{figure}
We have investigated the design of heterogeneous processors sharing a common ISA. The underlying motivation for single-ISA heterogeneity is that a diverse set of cores can enable runtime flexibility. We argue that selecting a diverse set of heterogeneous cores to enable flexible operation at runtime is a non-trivial problem due to diversity in program behaviour. We further show that common evaluation methods lead to false conclusions about diversity. We suggest the Kolmogorov--Smirnov (KS) test statistical test as an evaluation metric. The KS test is the first step towards a heterogeneous design methodology that optimises for runtime flexibility~\cite{Tomusk2014:PACT14,Tomusk2016:CAL16}.

A major roadblock to the further development of heterogeneous processors is the lack of appropriate evaluation metrics. Existing metrics can be used to evaluate individual cores, but to evaluate a heterogeneous processor, the cores must be considered as a collective. Without appropriate metrics, it is impossible to establish design goals for processors, and it is difficult to accurately compare two different heterogeneous processors. We present four new metrics to evaluate user-oriented aspects of sets of heterogeneous cores: localized non-uniformity, gap overhead, set overhead, and generality. The metrics consider sets rather than individual cores. We use examples to demonstrate each metric, and show that the metrics can be used to quantify intuitions about heterogeneous cores~\cite{Tomusk2015:TACO2015}.

For a heterogeneous processor to be effective, it must contain a diverse set of cores to match a range of runtime requirements and program behaviours. Selecting a diverse set of cores is, however, a non-trivial problem. We present a method of core selection that chooses cores at a range of power-performance points.
For example, we see on Fig.~\ref{fig:tomusk} that for a normalised power budget of 1.3 (1.3 times higher than the most power-efficient alternative), the best possible normalised time using the baseline selection is 1.75 (1.75 times the fastest execution time), whereas an 8 core selection can lower this ratio to 1.4 without exceeding the normalised power budget, i.e., our method brings a 20\% speedup.
Our algorithm is based on the observation that it is not necessary for a core to consistently have high performance or low power; one type of core can fulfil different roles for different types of programs. Given a power budget, cores selected with our method provide an average speedup of 7\% on EEMBC mobile benchmarks, and a 23\% on SPECint 2006 benchmarks over the state of the art core selection method~\cite{Tomusk2016:TACO16}.

\section{Holistic Optimisation Methods} \label{sec:holistic} 
\completedby{Imperial - Sajad}

\begin{figure}[t!]
\centering
\includegraphics[width = .85\linewidth]{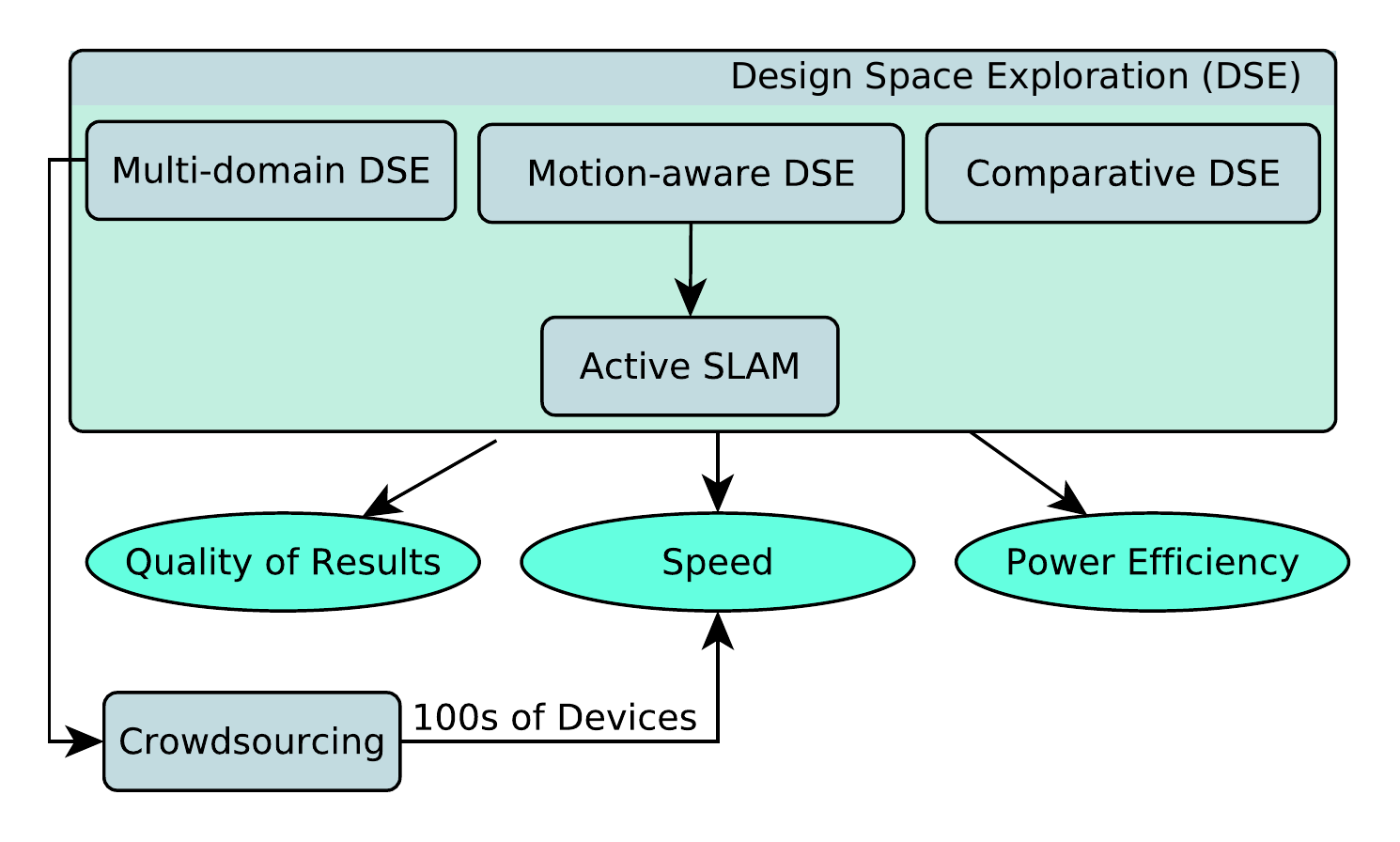}
\caption{Holistic optimisation methods explore all domains of the real-time 3D scene understanding, including hardware, software, and computer vision algorithms. Two holistic works presented here: \emph{Design Space Exploration} and \emph{Crowdsourcing}.}
\label{fig:ManHolistic}
\end{figure}

In this section, we introduce holistic optimisation methods that combine developments from multiple domains, i.e. hardware, software, and algorithm, to develop efficient end-to-end solutions.
The design space exploration work presents the idea of exploring many sets of possible parameters to properly exploit them at different situations.
The crowdsourcing further tests the DSE idea on a massive number of devices. Fig.~\ref{fig:ManHolistic} summarises their goals and contributions.

\subsection{Design Space Exploration} \completedby{Imperial - Sajad}

Design space exploration is the exploration of various possible design choices before running the system~\cite{Kang2011}. In scene understanding algorithms, the possible space of the design choices is very large and spans from high-level algorithmic choices down to parametric choices within an algorithm. 
For instance, Zhang et al.~\cite{Zhang2017RSS} explore algorithmic choices for a visual-inertial algorithmic parameters on an ARM CPU, as well as a Xilinx Kintex-7 XC7K355T FPGA. 
In this section, we introduce two DSE algorithms: The first one called multi-domain DSE explores algorithmic, compiler and hardware parameters. The second one, coined motion-aware DSE, further adds the complexity of the motion and the environment to the exploration space. The latter work is extended to develop an active SLAM algorithm.

\subsubsection{Multi-domain DSE}\label{sec:multidse} \completedby{Edinburgh - Bruno}

\begin{figure}[t!]
\centering
\includegraphics[width =.95 \linewidth]{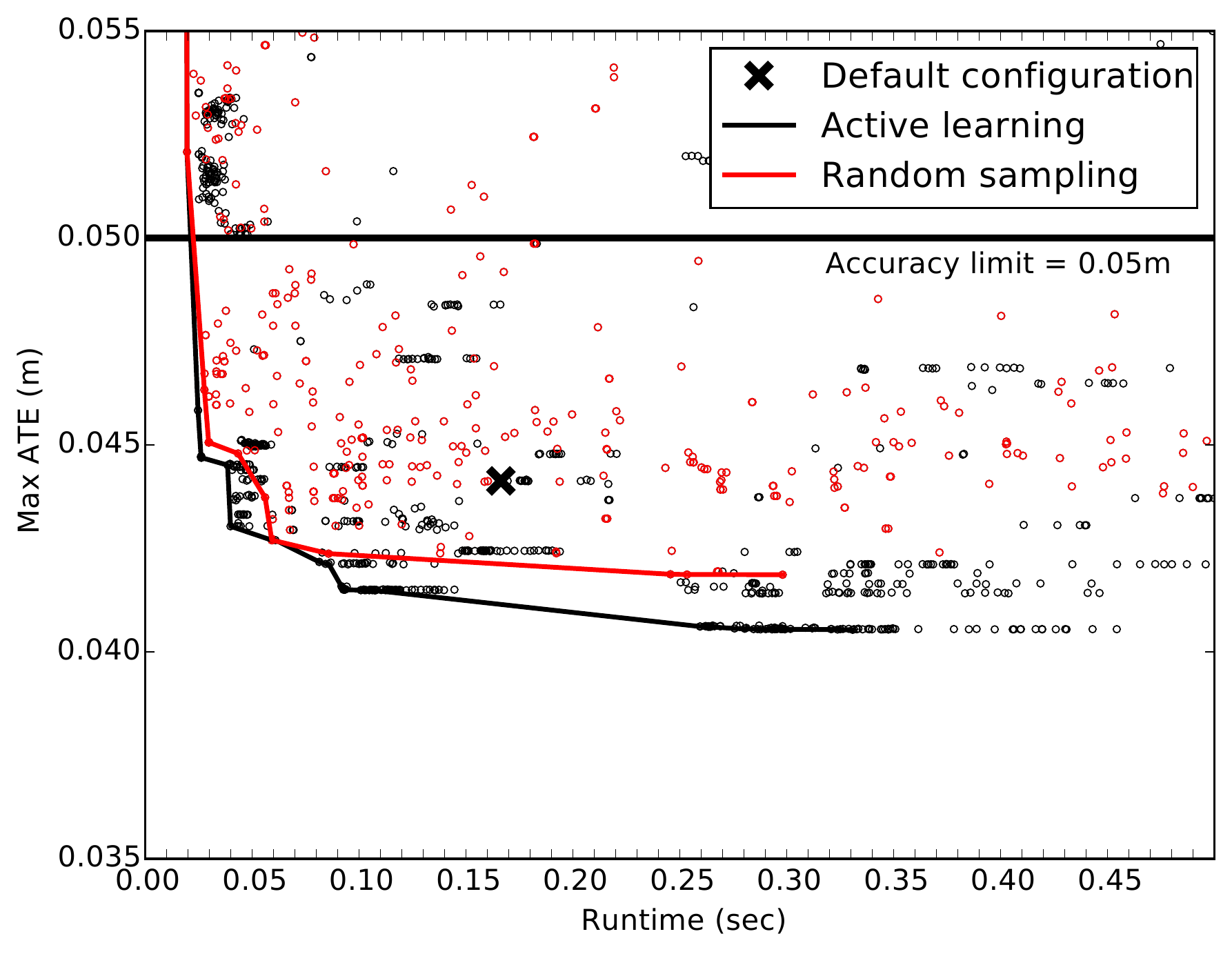}
\caption{
This plot illustrates the result of HyperMapper on the Design Space Exploration of the KinectFusion algorithmic parameters considering accuracy and frame rate metrics. We can see the result of random sampling (red) as well as the improvement of solutions after active learning (black).
}
\label{fig:dse}
\end{figure}

Until now, resource-intensive scene understanding algorithms, such as KinectFusion, could only run in real-time on powerful desktop GPUs. In~\cite{Bodin2016:PACT16} we examine
how it can be mapped to power constrained embedded systems and we introduce HyperMapper, a tool for multi-objective DSE. HyperMapper was demonstrated in a variety of applications ranging from computer vision and robotics to compilers~\cite{Bodin2016:PACT16,NardiBSVDK17,Saeedi_ICRA_2017,koeplinger2018spatial}. Key to our approach is the idea of incremental co-design
exploration, where optimisation choices that concern the domain layer are incrementally explored together with low-level compiler and
architecture choices (See Fig.~\ref{fig:activeslam}, dashed boxes).  The goal of this exploration is to reduce execution time while minimising power and meeting our quality of result objective. Fig.~\ref{fig:dse} shows an example performed with KinectFusion, in which for each point, a set of parameters, two metrics, maximum ATE and runtime speed, is shown.
As the design space is too large to exhaustively evaluate, we use active learning based on a random forest predictor to find good
designs.  We show that our approach can, for the first time, achieve dense 3D mapping and tracking in the real-time range 
within a 1W power budget on a popular embedded device. This is a 4.8x execution time improvement and a 2.8x power reduction compared to the state-of-the-art.

\subsubsection{Motion and Structure-aware DSE}\label{sec:msdse}  \completedby{Imperial - Sajad} 

\begin{figure}[t!]
\centering
\includegraphics[width = .85\linewidth]{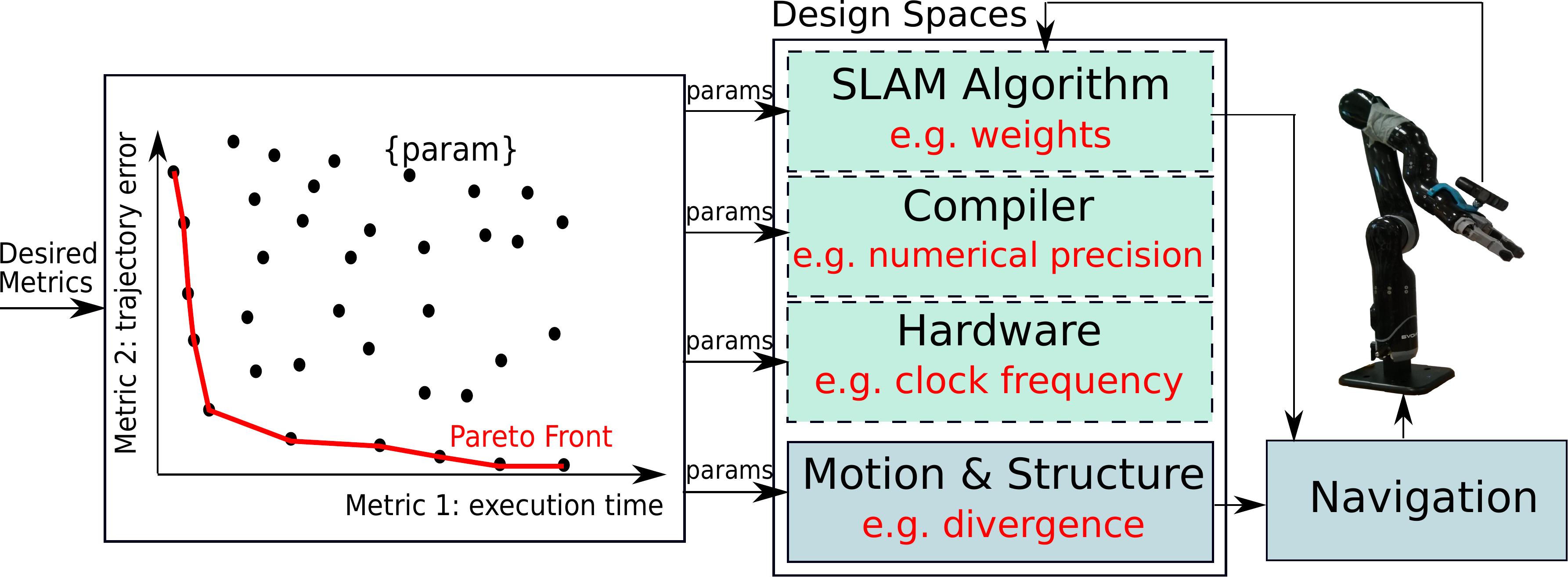}
\caption{Motion and structure aware active SLAM design space exploration using HyperMapper.}
\label{fig:activeslam}
\end{figure}

In Multi-domain DSE, when tuning software and hardware parameters, we also need to take into account the structure of the environment and the motion of the camera. In the Motion and Structure-aware Design Space Exploration (MS-DSE) work~\cite{Saeedi_ICRA_2017}, we determine the complexity of the structure and motion with a few parameters calculated using information theory. Depending on this complexity and the desired performance metrics, suitable parameters are explored and determined. The hypothesis of MS-DSE is that we can use a small set of parameters as a very useful proxy for a full description of the setting and motion of a SLAM application. We call these Motion and Structure (MS) parameters, and define them based on information divergence metric. Fig.~\ref{fig:activeslam} demonstrates the set of all design spaces.


MS-DSE presents a comprehensive parametrisation of 3D understanding scene algorithms, and thus based on this new parameterisation, many new concepts and applications can be developed. One of these applications, active SLAM, is outlined here. For more applications, please see~\cite{Bodin2016:PACT16,NardiBSVDK17,Saeedi_ICRA_2017,koeplinger2018spatial}.
\paragraph{Active SLAM}  \label{sec:activeslam} \completedby{Imperial - Sajad} 

\begin{figure}[t]
 \centering
 \includegraphics[width = 0.75\linewidth]{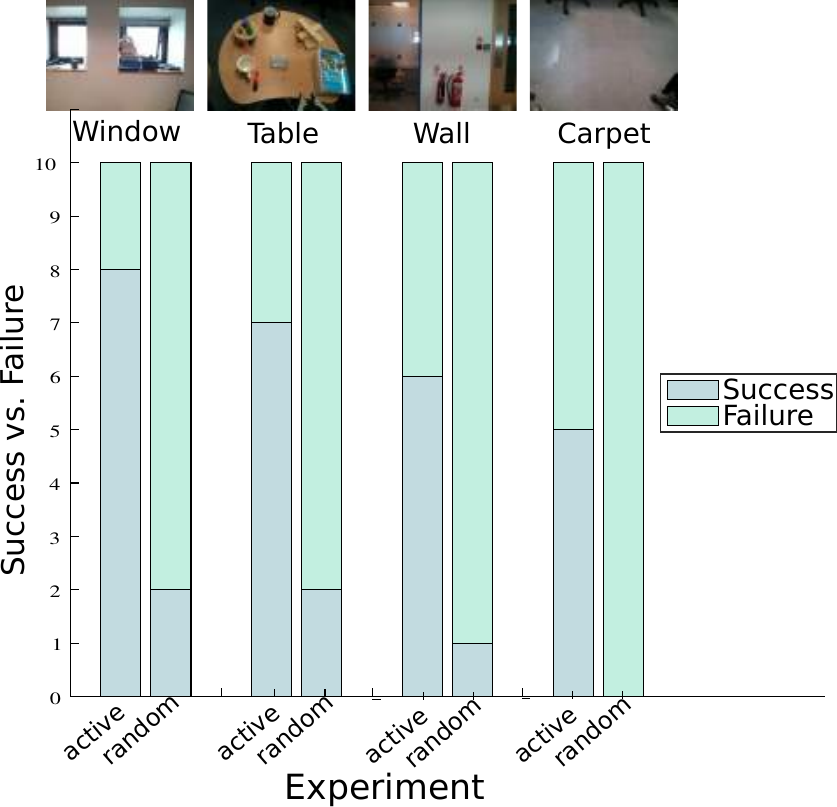}
    \caption{Success vs. failure rate when mapping the same environment with different motion planning algorithms: active SLAM and random walk.}
    \label{fig:activeslam_arm}
 \end{figure}

Active SLAM is the method for choosing the optimal camera trajectory, in order to maximise the camera pose estimation, the accuracy of the reconstruction, or the coverage of the environment. In~\cite{Saeedi_ICRA_2017}, it is shown that MS-DSE can be  utilised to optimise not only fixed system parameters, but also to guide a robotic platform to maintain a good performance for localisation and mapping. As shown in Fig.~\ref{fig:activeslam}, a Pareto front holds all optimal parameters. The front has been prepared in advance by exploring the set of all parameters. When the system is operating, optimal parameters are chosen given the desired performance metrics. Then these parameters are used to initialise the system. Using MS parameters, the objective is to avoid motions that cause very high statistical divergence between two consecutive frames. This way, we can provide a robust SLAM algorithm by allowing the tracking work all the time. 
Fig.~\ref{fig:activeslam_arm} compares the active SLAM with a random walk algorithm. 
The experiments were done in four different environments. In each environment, each algorithm was run 10 times. Repeated experiments serve as a measure of the robustness of the algorithm in dealing with uncertainties rising from minor changes in illumination, or inaccuracies of the response of the controller or actuator to the commands.
The consistency of the generated map was evaluated manually as either a success or failure of SLAM. If duplicates of one object were present in the map, it was considered as failure. 
This experiment shows more than 50~$\%$ success rate in SLAM when employing the proposed active SLAM algorithm~\cite{Saeedi_ICRA_2017}, an improvement in the \emph{robustness} of SLAM algorithms by relying on design space exploration.

\subsubsection{Comparative DSE of Dense vs Semi-dense SLAM} \label{sec:comparativedse}
\completedby{Imperial - Emanuele}

Another different direction in any DSE work is the performance exploration across multiple algorithms. While Multi-domain DSE explores different parameters of a given algorithm, the comparative DSE, presented in~\cite{zia2016comparative}, explores the performance of two different algorithms under different parametric choices.

In comparative DSE, two state-of-the-art SLAM algorithms, KinectFusion and LSD-SLAM, are compared on multiple datasets. Using SLAMBench benchmarking capabilities, a full design space exploration is performed over algorithmic parameters, compilation flags and multiple architectures. Such thorough parameter space exploration gives us key insights on the behaviour of each algorithm in different operative conditions and the relationship between different sets of distinct, yet correlated, parameters blocks. 

\begin{figure}[t!]
\centering
\includegraphics[width = \linewidth]{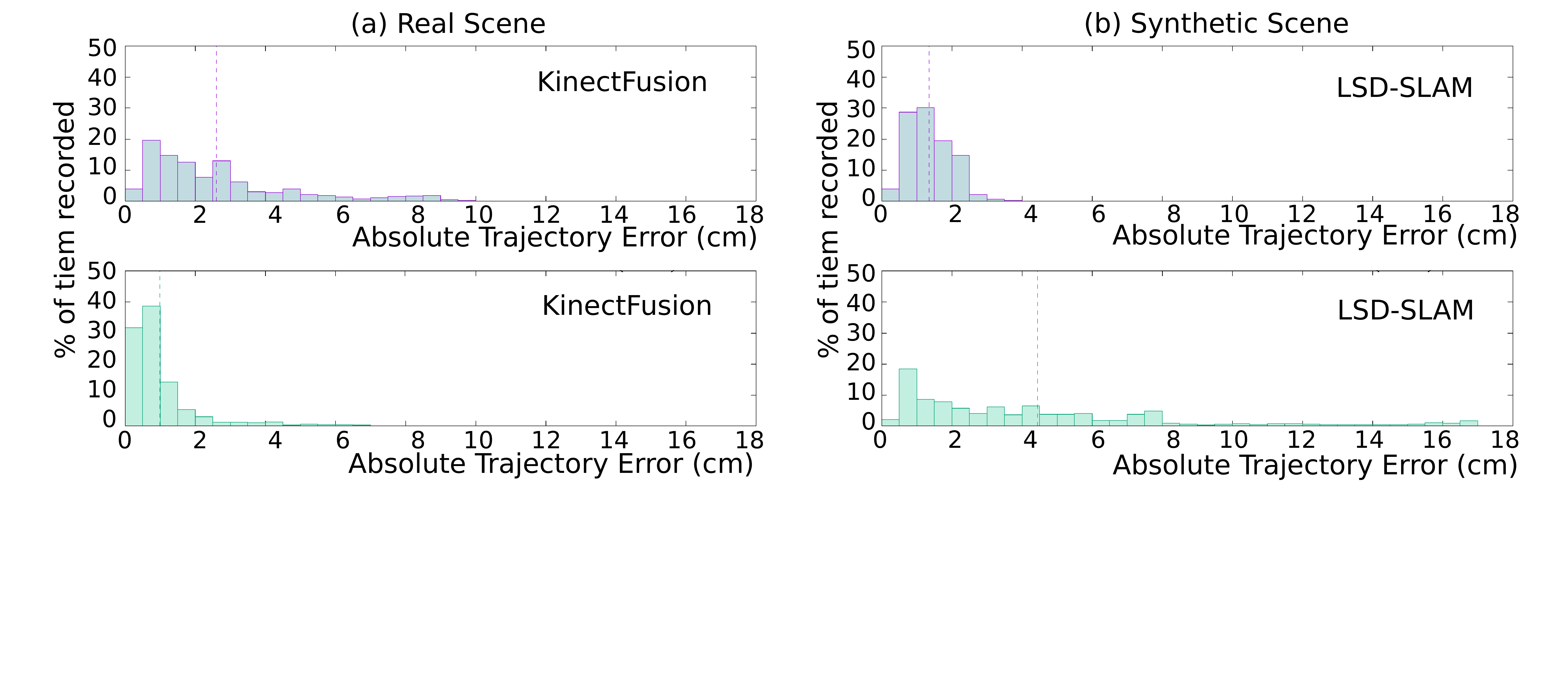}
\vspace{-15 mm}
\caption{Distribution of Absolute Trajectory Error (ATE) using KinectFusion and LSD-SLAM, run with default parameters on Desktop. The mean absolute error has been highlighted. (a) TUM RGB-D fr2\_xyz (b) ICL-NUIM lr\_kt2.}
\label{fig:lsd-kfusion-ate-dist}
\end{figure}
As an example, in Fig.~\ref{fig:lsd-kfusion-ate-dist} we show the result of comparative DSE between LSD-SLAM and KinectFusion in terms of their ATE distribution across two scenes of two different datasets. The histograms display the error distribution across the entire sequence, from which we can get a sense of how well the algorithms are performing for the \emph{whole} trajectory. We hope that these analyses enable researchers to develop more robust algorithms. Without the holistic approach enabled by SLAMBench such insights would have been much harder to obtain. This sort of information is invaluable for a wild range of SLAM practitioners, from VR/AR designers to roboticists that want to select/modify the best algorithm for their particular use case.

\subsection{Crowdsourcing} \label{sec:crowdsourcing}
\completedby{Edinburgh - Bruno}

The SLAMBench framework and more specifically its various KinectFusion implementations has been ported to Android. More than 2000 downloads have been made since its official release on the Google Play store. We received numerous positive feedback reports and this application has generated a great deal of interest in the community and with industrial partners.

\begin{figure}[t!]
\centering
\includegraphics[width =.85 \linewidth]{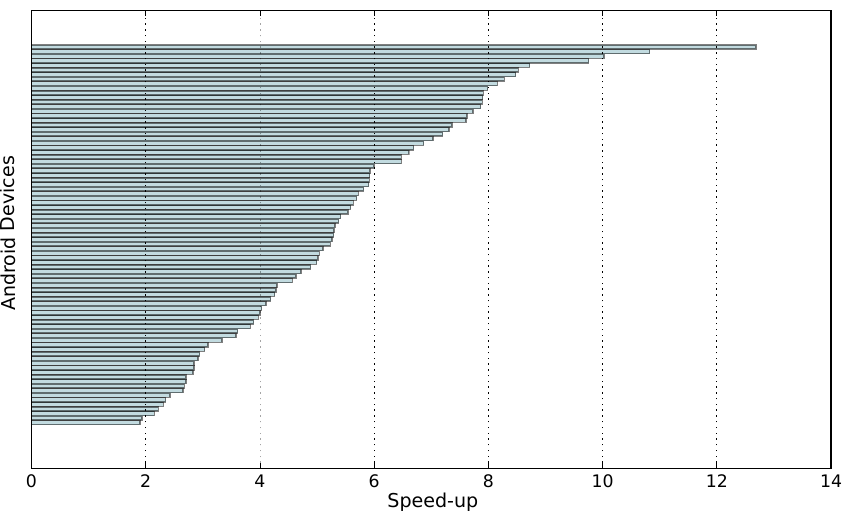}
\caption{By combining design space exploration and crowdsourcing, we checked that design space exploration efficiently works on various types of platforms. This figure demonstrates the speed-up of the KinectFusion algorithm on various different types of Android devices. Each bar represents the speed-up for one type (model) of Android device.  The models are not shown for the sake of clarity of the figure. 
} 
\label{fig:crownsource}
\end{figure}

This level of uptake allowed us to collect data from more than 100 different mobile phones
Fig.~\ref{fig:crownsource} shows the speed-up across many models of Android devices that we have experimented with. Clearly it is possible to achieve more than twice \emph{runtime speed} by tuning the system parameters using the tools introduced in the paper.
We plan to use these data to analyse the performance of KinectFusion on those platforms, and to provide techniques to optimise KinectFusion performance depending of the targeted platform. This work will apply transfer-learning methodology. We believe that by combining design space exploration~\cite{NardiBSVDK17} and the collected data, we can train a decision machine to select code variants and configurations for diverse mobile platforms automatically.

\section{Conclusion} \label{sec:conclusions}


In this paper we focused on SLAM, which is an enabling technology in many fields including virtual reality, augmented reality, and robotics. The paper presented several contributions across hardware architecture, compiler and runtime software systems, and computer vision algorithmic layers of SLAM pipeline. We proposed not only contributions at each layer, but also holistic methods that optimise the system as a whole.

In computer vision and applications, we presented benchmarking tools that allow us to select a proper dataset and use it to evaluate different SLAM algorithms. SLAMBench is used to evaluate the KinectFusion algorithm on various different hardware platforms. SLAMBench2 is used to compare various SLAM algorithms very efficiently. 
We also extended the KinectFusion algorithm, such that it can be used in robotic path planning and navigation algorithms by mapping both occupied and free space of the environment. Moreover, we explored new sensing technologies such as focal-plane sensor-processor arrays, which have low power consumption and high effective frame rate.

The software layer of this project demonstrated that software optimisation can be used to deliver significant improvements in power consumption and speed trade-off when specialised for computer vision applications. We explored {\em static}, {\em dynamic}, and {\em hybrid} approaches and focused their application on the KinectFusion algorithm. 
Being able to select and deploy optimisations adaptively is particularly beneficial in the context of dynamic runtime environment where application-specific details can strongly improve the result of JIT compilation and thus the speed of the program.

The project has made a range of contributions across the hardware design and development field. Profiling tools have been developed in order to locate and evaluate performance bottlenecks in both native and managed applications. These bottlenecks could then be addressed by a range of specialisation techniques, and the specialised hardware evaluated using the presented simulation techniques. This represents a full workflow for creating new hardware for computer vision applications which might be used in future platforms.

Finally, we report on holistic methods that exploit our ability to explore the design space at every level in a holistic fashion. We demonstrated several design space exploration methods where we showed that it is possible to fine-tune the system such that we can meet desired performance metrics. It is also shown that we can increase public engagement in accelerating the design space exploration by crowdsourcing.

In future work, two main directions will be followed: The first is exploiting our knowledge from all domains of this paper to select a SLAM algorithm and design a chip that is customised to efficiently implement the algorithm. 
This approach will utilise data from SLAMBench2 and real-world experiments to drive the design of a specialised vision processor. The second direction is utilising the tools and techniques presented here to develop a standardised method that takes the high-level scene understanding functionalities and develops the optimal code that maps the functionalities to the heterogeneous resources available, optimising for the desired performance metrics.

\section{Acknowledgements}
This research is supported by Engineering and Physical Sciences Research Council (EPSRC), grant reference EP/K008730/1, PAMELA project.

\bibliographystyle{IEEEtran}
\bibliography{vision}

\end{document}